\documentclass{ieeeaccess}

\usepackage{cite}
\usepackage{amsmath,amssymb,amsfonts}
\usepackage{algorithmic}
\usepackage{graphicx}
\usepackage{multirow}
\usepackage{textcomp}
\usepackage{url}
\usepackage[nolist]{acronym}
\usepackage[inline]{enumitem}
\usepackage{siunitx}
\usepackage{hyperref}
\usepackage{csquotes}

\usepackage{bm}
\makeatletter
\AtBeginDocument{\DeclareMathVersion{bold}
\SetSymbolFont{operators}{bold}{T1}{times}{b}{n}
\SetSymbolFont{NewLetters}{bold}{T1}{times}{b}{it}
\SetMathAlphabet{\mathrm}{bold}{T1}{times}{b}{n}
\SetMathAlphabet{\mathit}{bold}{T1}{times}{b}{it}
\SetMathAlphabet{\mathbf}{bold}{T1}{times}{b}{n}
\SetMathAlphabet{\mathtt}{bold}{OT1}{pcr}{b}{n}
\SetSymbolFont{symbols}{bold}{OMS}{cmsy}{b}{n}
\renewcommand\boldmath{\@nomath\boldmath\mathversion{bold}}}
\makeatother

\def\BibTeX{{\rm B\kern-.05em{\sc i\kern-.025em b}\kern-.08em
    T\kern-.1667em\lower.7ex\hbox{E}\kern-.125emX}}

\newacro{tof}[ToF]{Time-of-Flight}
\newacro{ai}[AI]{Artifical Inteligence}
\newacro{gpu}[GPU]{Graphical Processing Unit}
\newacro{api}[API]{Application Programming Interface}
\newacro{slam}[SLAM]{Simultaneous Localization and Mapping}
\newacro{fps}[FPS]{Frames per Second}
\newacro{fov}[FOV]{Field of View}
\newacro{icp}[ICP]{Iterative Closest Point}
\newacro{tp}[TP]{True Positive}
\newacro{fp}[FP]{False Positive}
\newacro{fn}[FN]{False Negative}
\newacro{sd}[SD]{Standard Deviation}
\newacro{cd}[CD]{Chamfer Distance}
\newacro{js}[JS]{Jaccard Similarity}
\newacro{sift}[SIFT]{Scale-Invariant Feature Transform}
\newacro{sfm}[SfM]{Structure from Motion}

\newcommand{\etal}[1]{#1 \textit{et al.}}

\newcommand{\equationref}[1]{\hyperref[#1]{Eq.~\ref*{#1}}}
\newcommand{\figref}[1]{\hyperref[#1]{Fig.~\ref*{#1}}}
\newcommand{\tabref}[1]{\hyperref[#1]{Table~\ref*{#1}}}
\newcommand{\secref}[1]{\hyperref[#1]{Section~\ref*{#1}}}
\newcommand{\algoref}[1]{\hyperref[#1]{Alg.~\ref*{#1}}}

\newcommand{\sotan}{state of the art}

\usepackage[firstpageonly=true]{draftwatermark}

\SetWatermarkAngle{0}
\SetWatermarkColor{black}
\SetWatermarkLightness{0.5}
\SetWatermarkFontSize{9pt}
\SetWatermarkVerCenter{20pt}
\SetWatermarkHorCenter{0.65\textwidth}
\SetWatermarkText{\parbox{30cm}{%
\centering This is the final version of the manuscript accepted to IEEE Access. \\
\centering See the published version at: \url{https://doi.org/10.1109/ACCESS.2025.3560810}\\
\centering (c) 2025 The Authors. This work is licensed under a Creative Commons Attribution 4.0 License
}}
\begin{document}
\history{Received 10 March 2025, accepted 2 April 2025, date of publication 15 April 2025, date of current version 24 April 2025.}
\doi{10.1109/ACCESS.2025.3560810}

\title{Empirical Comparison of Four Stereoscopic Depth Sensing Cameras for Robotics Applications}

\author{\uppercase{Lukas Rustler}\authorrefmark{1, *},
\uppercase{Vojtech Volprecht}\authorrefmark{1, *}, and \uppercase{Matej Hoffmann}\authorrefmark{1}}

\def\mainauthor{Rustler}

\address[1]{Department of Cybernetics, Faculty of Electrical Engineering, Czech Technical University in Prague}
\address[*]{The authors contributed equally}

\tfootnote{This work was supported by the Czech Science Foundation (GACR), project no. 25-18113S and co-funded by the European Union under the project Robotics and Advanced Industrial Production (reg. no. CZ.02.01.01/00/22\_008/0004590). L.R. was additionally supported by the Grant Agency of the Czech Technical University in Prague, grant No. SGS24/096/OHK3/2T/13.}

\makeatletter
\markboth
{\mainauthor{} \headeretal: \@title}
{\mainauthor{} \headeretal: \@title}
\makeatother

\corresp{Corresponding author: Matej Hoffmann (e-mail: matej.hoffmann@fel.cvut.cz).}

%150-250 words; one paragraph;
\begin{abstract}
Depth sensing is an essential technology in robotics and many other fields. Many depth sensing (or RGB-D) cameras are available on the market and selecting the best one for your application can be challenging. In this work, we tested four stereoscopic RGB-D cameras that sense the distance by using two images from slightly different views. We empirically compared four cameras (Intel RealSense D435, Intel RealSense D455, StereoLabs ZED 2, and Luxonis OAK-D Pro) in three scenarios: (i) planar surface perception, (ii) plastic doll perception, (iii)  household object perception (YCB dataset). We recorded and evaluated more than 3,000 RGB-D frames for each camera. For table-top robotics scenarios with distance to objects up to one meter, the best performance is provided by the D435 camera that is able to perceive with an error under 1 cm in all of the tested scenarios. For longer distances, the other three models perform better, making them more suitable for some mobile robotics applications. OAK-D Pro additionally offers integrated AI modules (e.g., object and human keypoint detection). ZED 2 is overall the best camera which is able to keep the error under 3 cm even at 4 meters. However, it is not a standalone device and requires a computer with a GPU for depth data acquisition. All data (more than 12,000 RGB-D frames) are publicly available at https://rustlluk.github.io/rgbd-comparison. 
\end{abstract}

%key words of phrases; in alphabetical order
\begin{keywords}
Depth Sensing, Intel RealSense, Luxonis OAK-D Pro, ZED 2, Object Detection, RGB-D, Segmentation
\end{keywords}

\titlepgskip=-21pt

\maketitle

\section{Introduction}
\label{sec:introduction}
\PARstart{D}{epth} sensing cameras, or RGB-D cameras, are visual sensors that provide standard images (RGB), along with depth information. They are compact, powerful, and affordable sensors which make them attractive for a wide range of applications in robotics and many other fields. Depth information is useful when building a map of an environment, so RGB-D cameras are widely used in \ac{slam}~\cite{zhang2021survey,sturm2012benchmark} or scene reconstruction~\cite{wang2024MorpheuSNeuralDynamic, li2022HighqualityIndoorScene, azinovic2022NeuralRGBDSurface}. They can also be used for 3D reconstruction or shape completion of individual objects~\cite{yan2022ShapeFormerTransformerBasedShapeb, rustler2023efficient}. An important table-top task that utilizes depth sensing is grasping~\cite{gui2025graspfast, qin2023RGBDGraspDetection, shi2024ASGraspGeneralizableTransparent}. An interesting application is human pose and shape estimation~\cite{pesavento2024ANIMAccurateNeural, huang2024InterCapJointMarkerless, Docekal_closeProximity}. The output can then be used, for example, in human-robot interaction~\cite{secil2022MinimumDistanceCalculation, Docekal_closeProximity}.

In this work, we compare four different RGB-D cameras: Intel RealSense D435~\cite{DepthCameraD435}, Intel RealSense D455~\cite{DepthCameraD455}, StereoLabs ZED 2~\cite{ZED2iIndustrial}, and Luxonis OAK-D Pro~\cite{OAKDPro}. RGB-D cameras have many applications. However, currently the \sotan{} in camera comparison is usually focusing only on performance on flat surfaces, which is not representative of the different application scenarios. The goals of this work are thus two-fold. First, we want to complement the performance comparison on planar surfaces with the perception of 3D objects with complicated geometries, focusing also on shorter distances relevant for robot manipulation, for example. Second, the set of cameras we tested is new and aims to reflect the most relevant models available on the market now. 

All of the cameras selected in this work operate on the stereoscopic principle---the actual physical sensor contains two cameras with a fixed distance, and the depth is computed from matching features in the two streams. RealSense and OAK-D Pro devices are active, i.e., they project an IR grid that helps with feature matching. Depth can be also perceived using \ac{tof} sensors, like LIDAR, which perceive depth more accurately, but are more bulky and expensive. They are typically used on large mobile robots and are not deployed in indoor robot manipulation \cite{drouin2020ConsumerGradeRGBDCameras}.

The accuracy of the sensors is crucial for the correct use and deployment. However, manufacturers usually provide only technical specifications and information on the percentual accuracy in a given depth~\cite{DepthCameraD435, DepthCameraD455, OAKDPro, ZED2iIndustrial}. This information is useful, but it is not informative about the standard deviation in individual underlying tasks and was usually measured under unknown laboratory conditions that are hard to replicate in daily use. Researchers always wanted to better estimate the camera capabilities in computer vision applications, as done by \etal{Andersen}~\cite{andersen2012kinect} with the Kinect or \etal{Nguyen}~\cite{NguyenKinectNoise2012} who tried to estimate the noise of the Kinect to improve its performance in \ac{slam}. After Kinect, Intel cameras became popular. \etal{Carfagni} provided a performance evaluation of Intel RS300~\cite{carfagni2017PerformanceIntelSR300}. Some researchers also compared stereo vison cameras with \ac{tof} cameras~\cite{lourenco2021IntelRealSenseSR305} or only \ac{tof}~\cite{lopezparedes2023PerformanceEvaluationStateoftheArt}. \etal{Tadic} performed comparison of Intel and StereoLabs cameras from the point-of-view of technical specifications and from visual inspection of the depth images~\cite{tadicPerspectivesRealSenseZED2022}. Other ZED camera analysis was performed in \cite{abdelsalam2024DepthAccuracyAnalysis}, where the authors tested the camera in indoor environments. Halmetschlager et al.~\cite{halmetschlager-funekEmpiricalEvaluationTen2019} compared 10 depth-sensing cameras with different working principles, proposed a noise model, and advised which camera to use for which purpose.

Some works focused more on accuracy in a given task, e.g., bolt tightening~\cite{dias2023Comparison3DSensors} or fruit detection and sizing~\cite{fu2020ApplicationConsumerRGBDd, neupane2021EvaluationDepthCameras}. More unusual research containing RGB-D cameras can be the comparison with multi-view stereo for face surface reconstruction~\cite{chen2020ComparativeAnalysisActive} or learning how to recognize a camera model or light source only from depth noise~\cite{haider2022WhatCanWe}. Similar comparisons to ours were made by \etal{Heinemann}~\cite{heinemann2022MetrologicalApplicationrelatedComparison} who compared six cameras, including the ones we have, in bias and standard deviation on flat surfaces, spheres, and one 3D box. Other recent and comparable study was proposed by \etal{Servi}~\cite{servi2024ComparativeEvaluationIntel}. The authors compared 4 cameras (not the same set as us) using planes, spheres, and a 3D printed statue.

\textbf{Contribution.} The unique contribution of this work is an extensive experimental evaluation of four state-of-the-art RGB-D cameras on object perception. While previous works have mostly assessed performance on planar surfaces, we additionally evaluate performance on a plastic doll figure and on a standard robotics benchmark object dataset (YCB). Furthermore, we have employed five different evaluation metrics, including those used in 3D object reconstruction (Chamfer Distance and Jaccard Similarity). Our conclusions are thus directly relevant to the object perception and robot manipulation community. We provide an overview of the cameras' technical parameters as well as their user interfaces.
Finally, we make our dataset with more than 12,000 segmented RGB-D images publicly available for the community. This dataset can serve not only to estimate the camera precision, but also as ground truth for RGB-D segmentation or object perception pipelines.

\section{Cameras and Specifications}
\label{sec:cameras_and_specifications}

\begin{table*}[ht]
\caption{Technical parameters of the cameras used in this study. *Stated by the manufacturers. **Not with maximal resolution.}
\label{tab:camera_properties}
\centering
\resizebox{\textwidth}{!}{%
\begin{tabular}{ccc|cccc|ccc|ccc}
\cline{4-10}
 &  &  & \multicolumn{4}{c|}{\textbf{Depth}} & \multicolumn{3}{c|}{\textbf{RGB}} &  &  &  \\ \hline
\multicolumn{1}{|c|}{\multirow{2}{*}{\textbf{Camera}}} & \multicolumn{1}{c|}{\textbf{Stereo}} & \textbf{Ideal} & \multicolumn{1}{c|}{\multirow{2}{*}{\textbf{Accuracy*}}} & \multicolumn{1}{c|}{\multirow{2}{*}{\textbf{FOV}}} & \multicolumn{1}{c|}{\textbf{Max}} & \textbf{Max} & \multicolumn{1}{c|}{\multirow{2}{*}{\textbf{FOV}}} & \multicolumn{1}{c|}{\textbf{Max}} & \textbf{Max} & \multicolumn{1}{c|}{\multirow{2}{*}{\textbf{Dimensions [mm]}}} & \multicolumn{1}{c|}{\multirow{2}{*}{\textbf{GPU}}} & \multicolumn{1}{c|}{\multirow{2}{*}{\textbf{AI}}} \\
\multicolumn{1}{|c|}{} & \multicolumn{1}{c|}{\textbf{Technology}} & \textbf{Range [m]} & \multicolumn{1}{c|}{} & \multicolumn{1}{c|}{} & \multicolumn{1}{c|}{\textbf{Resolution}} & \textbf{FPS} & \multicolumn{1}{c|}{} & \multicolumn{1}{c|}{\textbf{Resolution}} & \textbf{FPS} & \multicolumn{1}{c|}{} & \multicolumn{1}{c|}{} & \multicolumn{1}{c|}{} \\ \hline
\multicolumn{1}{|c|}{\textbf{D435}} & \multicolumn{1}{c|}{Active} & 0.3 - 3 & \multicolumn{1}{c|}{$<$2\% at 2\;m} & \multicolumn{1}{c|}{$87^{\circ} \times 58^{\circ}$} & \multicolumn{1}{c|}{1280x720} & 90** & \multicolumn{1}{c|}{$69^{\circ} \times 42^{\circ}$} & \multicolumn{1}{c|}{1920x1080} & 30 & \multicolumn{1}{c|}{90x25x25} & \multicolumn{1}{c|}{$\boldsymbol{\times}$} & \multicolumn{1}{c|}{$\boldsymbol{\times}$} \\ \hline
\multicolumn{1}{|c|}{\textbf{D455}} & \multicolumn{1}{c|}{Active} & 0.6 - 6 & \multicolumn{1}{c|}{$<$2\% at 4\;m} & \multicolumn{1}{c|}{$87^{\circ} \times 58^{\circ}$} & \multicolumn{1}{c|}{1280x720} & 90** & \multicolumn{1}{c|}{$90^{\circ} \times 65^{\circ}$} & \multicolumn{1}{c|}{1280x800} & 30 & \multicolumn{1}{c|}{124x26x29} & \multicolumn{1}{c|}{$\boldsymbol{\times}$} & \multicolumn{1}{c|}{$\boldsymbol{\times}$} \\ \hline
\multicolumn{1}{|c|}{\textbf{ZED 2}} & \multicolumn{1}{c|}{Passive} & 0.3 - 20 & \multicolumn{1}{c|}{$<$0.8\% at 2\;m} & \multicolumn{1}{c|}{$110^{\circ} \times 70^{\circ}$} & \multicolumn{1}{c|}{2208x1242} & 100** & \multicolumn{1}{c|}{$110^{\circ} \times 70^{\circ}$} & \multicolumn{1}{c|}{2208x1242} & 100** & \multicolumn{1}{c|}{175x30x32} & \multicolumn{1}{c|}{$\checkmark$} & \multicolumn{1}{c|}{$\checkmark$} \\ \hline
\multicolumn{1}{|c|}{\textbf{OAK-D Pro}} & \multicolumn{1}{c|}{Active} & 0.8 - 12 & \multicolumn{1}{c|}{$<$2\% at 4\;m} & \multicolumn{1}{c|}{$80^{\circ} \times 55^{\circ}$} & \multicolumn{1}{c|}{1280x800} & 120 & \multicolumn{1}{c|}{$66^{\circ} \times 54^{\circ}$} & \multicolumn{1}{c|}{4056x3040} & 60 & \multicolumn{1}{c|}{97x23x30} & \multicolumn{1}{c|}{$\boldsymbol{\times}$} & \multicolumn{1}{c|}{$\checkmark$} \\ \hline
\end{tabular}%
}
\end{table*}

\begin{figure}[htb]
    \centering
    \begin{minipage}[t]{0.49\columnwidth}
        \centering
        \includegraphics[width=0.514\textwidth]{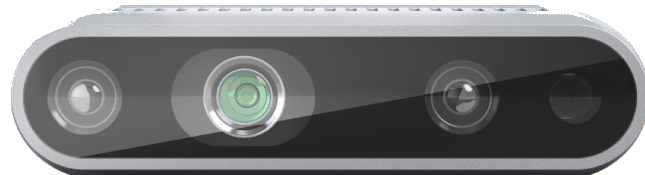}\\
        (a) RealSense D435.
    \end{minipage}
    \begin{minipage}[t]{0.49\columnwidth}
        \centering
        \includegraphics[width=0.708\textwidth]{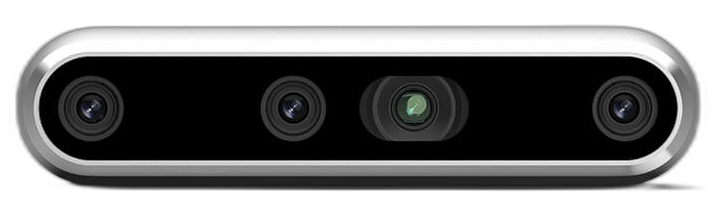}\\
        (b) RealSense D455.
    \end{minipage}\\
    \begin{minipage}[t]{0.49\columnwidth}
        \centering
        \includegraphics[width=1\textwidth]{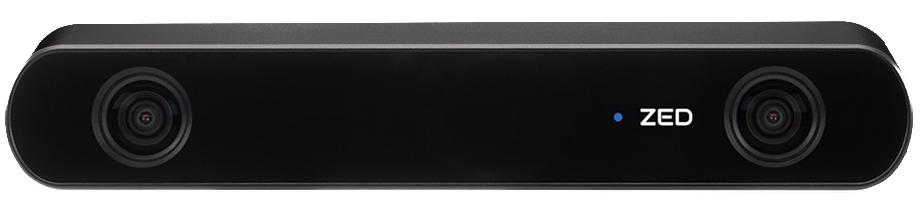}\\
        (c) StereoLabs ZED 2.
    \end{minipage}
    \begin{minipage}[t]{0.49\columnwidth}
        \centering
        \includegraphics[width=0.554\textwidth]{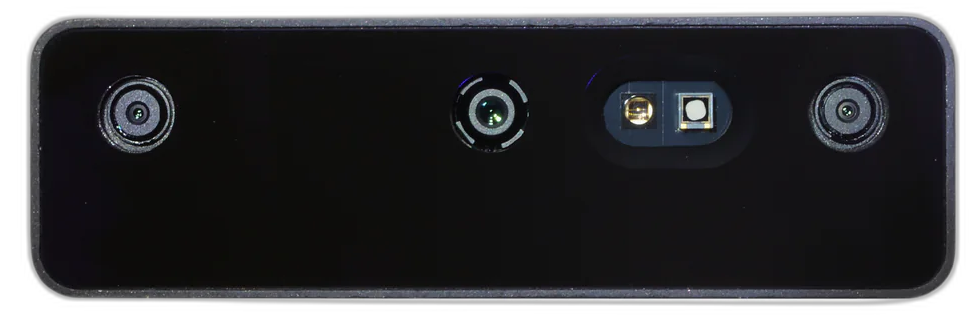}\\
        (d) Luxonis OAK-D Pro.
    \end{minipage}
    \caption{RGB-D cameras used in the experiments. The cameras are shown in scale---width of D435 camera in a) is 90\;mm.}
    \label{fig:cameras}
\end{figure}

This section describes the hardware used and its technical specifications. The RGB-D cameras used in this work can be seen in \figref{fig:cameras} and their main technical specifications can be found in \tabref{tab:camera_properties}. The selection was motivated by an informal survey through the recent literature and among colleagues in robotics (mobile robotics and robot manipulation). All four cameras use the stereoscopic effect to compute the depth. In other words, the depth is computed from correspondences in two images from slightly different views. Both RealSense devices and the OAK-D Pro camera use active stereo, i.e., they project additional IR points that help to find correspondences. 

\textbf{On-board computation.} The Intel cameras (D435, D455) perform depth computation on-board, and the resources of a host computer are used only for potential post-processing. ZED 2, on the other hand, requires the host computer to contain CUDA-enabled \ac{gpu} to compute the depth. The benefit arising from \ac{gpu} necessity is that StereoLabs provides a wide selection of \ac{ai} tools, such as object detection or body tracking, which can be run through their \ac{api} directly using camera output. The OAK-D Pro camera performs all computations on board, but still provides \ac{ai} features that also run directly on the camera. 

\textbf{Depth accuracy.} In terms of depth accuracy reported by the respective manufacturers, D455, ZED 2, and OAK-D Pro should perform similarly with an error of $<$ 2\% at 4\;m. The D435 should, according to the specifications, be twice less accurate with error of $<$2\% at 2\;m. The devices that do not require \ac{gpu} (D435, D455, OAK-D Pro) have a similar maximal depth resolution (1280x720). However, the \ac{fps} with the maximal resolution is 30 for D435, D455 and 120 for OAK-D Pro. RealSense devices can get up to 90, but the resolution needs to be lower. These three cameras also have a similar \ac{fov}. For ZED 2, the maximum resolution can be up to 2K with 15\ac{fps} (ZED 2 can get up to 100\ac{fps} based on the resolution) with larger \ac{fov}.

\textbf{Dimensions and range.} The cameras also differ in their dimensions. D435 and OAK-D Pro are both quite small (less than 10\;cm in length), D455 is longer, and ZED 2 is the longest, with almost twice the length as for D435. The physical dimension results in a different stereo baseline and influences the range of the sensor. The ideal range for 435 is [0.3\;m, 3\;m], whereas its [0.6\;m, 6\;m] for D455, [0.8\;m, 12\;m] for OAK-D Pro (this is an exception, as the range is enhanced with custom processing) and [0.3\;m, 20\;m] for ZED 2. However, the D435 with the shortest stereo baseline and limited maximum range is, on the other hand, best suited for depth perception close to the camera. 

\textbf{Price.} RealSense D435 costs 314\$, Luxonis OAK-D Pro 349\$, RealSense D455 419\$ and StereoLabs ZED2 449\$. Prices are from the respective official websites at the time of writing this article.

\section{Setup and Evaluation}
\label{sec:setup_and_metrics}

\begin{figure}[htb]
    \centering
    \includegraphics[width=1\columnwidth]{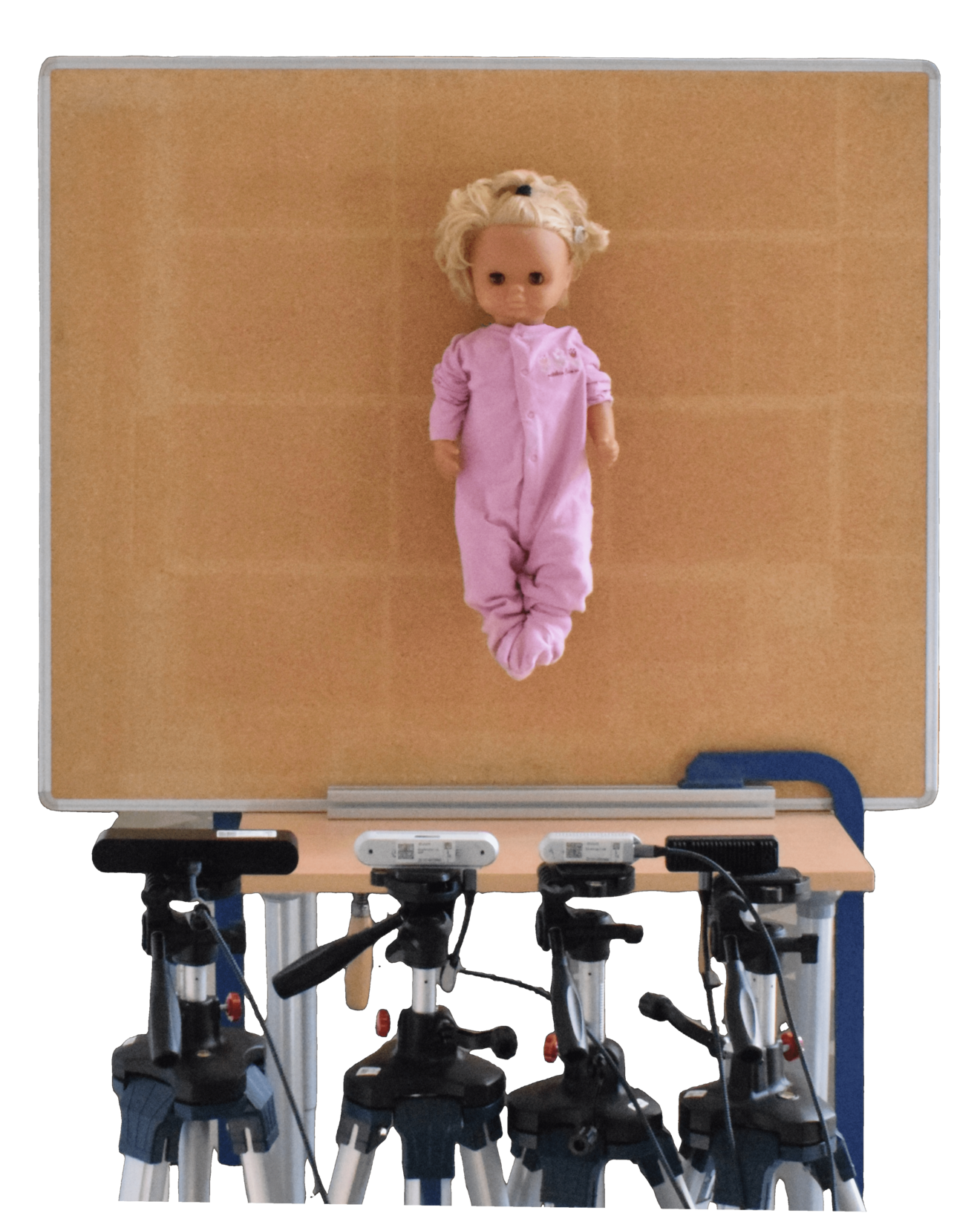}
    \caption{Experimental setup illustration -- plastic doll perception.}
    \label{fig:setup}
\end{figure}

The setup utilized to record the experiments is depicted in \figref{fig:setup}. We mounted the four cameras on tripods in a long hallway and set the tripods so that the sensors of the individual sensors were at the same height and level. The distance was measured using a laser measuring device from a plane surface. The same height was achieved using a long spirit level. In front of the cameras, there is a table. We always moved the table or objects on top of the table, not the cameras. We tested three scenarios: 
\begin{enumerate*}[label=(\roman*)]
    \item planar surface perception, where only the planar surface behind the doll in \figref{fig:setup} was recorded;
    \item plastic doll perception, where a doll was captured by the cameras (corresponds to \figref{fig:setup});
    \item and YCB objects perception, where objects from the YCB Object and Model set ~\cite{ycb_dataset} \url{https://www.ycbbenchmarks.com/} were perceived (without the planar surface behind).
\end{enumerate*}

The scenarios were selected to compare the cameras in terms of flat surface sensing (planes) and on more complicated objects (doll, YCB). Flat surfaces are important in mobile robotics (e.g., navigation in indoor environments) or to easily segment tables in table-top scenarios. The plastic doll was selected as a complicated 3D object and at the same time a human figure, as RGB-D cameras are also used in human pose estimation and shape reconstruction scenarios. The YCB dataset is widely used for benchmarking in object segmentation and robot manipulation; the video variant YCB-Video~\cite{Xiang2018-posecnn} is used in 6D pose estimation.

The distances from the camera to the objects were measured with a laser measuring device and the data were captured under constant light conditions of about 420\;lx. As we recorded in a long hallway, we did not have a proper way to control the light conditions. Thus, to keep the comparison as fair and as consistent as possible, we switched on the light (which were equally distributed over the distance) and checked the light intensity before recording at every distance using a lux meter.

The cameras were connected to one computer with a NVIDIA RTX 3070 \ac{gpu} and Intel Core i9-11900 CPU. To eliminate computational processing requirements, we always recorded the output from one camera at a time. RealSense cameras were recorded through the RealSense Viewer application. ZED 2 and OAK-D Pro through their respective Python \ac{api}. We used default settings for each camera---see \tabref{tab:camera_settings}. No post-processing was applied.

\begin{table}[htb]
\caption{Camera settings used in the experiments. *Name in the corresponding \ac{api}. All are default for the given camera.}
\label{tab:camera_settings}
\centering
\begin{tabular}{|c|c|c|c|}
\hline
\textbf{Camera} & \textbf{Depth Resolution} & \textbf{FPS} & \textbf{Mode*} \\ \hline
\textbf{D435} & 848x480 & 30 & Default \\ \hline
\textbf{D455} & 848x480 & 30 & Default \\ \hline
\textbf{ZED 2} & 1920x1080 & 30 & ULTRA \\ \hline
\textbf{OAK-D Pro} & 1280x800 & 30 & HIGH\_ACCURACY \\ \hline
\end{tabular}
\end{table}

\subsection{Data Acquisition and Processing}
In all scenarios, we always recorded the static scene for at least two seconds with every camera. However, the scenarios require different data processing for proper evaluation.

\begin{itemize}
    \item Planes -- In this case, we only moved the table further away from the cameras to different distances. Then, from the resulting point cloud, we extracted a square with a side of 10\;cm with a center in front of the camera. 

    \item Plastic doll and YCB objects -- In this case, we moved the table further away and also moved the individual objects so that they are in front of the respective camera all the time. Then, from the resulting point clouds, we segmented the object. We cropped the point clouds in such a way to keep only a region around the distance at which we recorded the object. At lower distances, we further utilized the RANSAC~\cite{Fischer81_RANSAC} algorithm to automatically extract the largest plane---the table---and cropped everything \enquote{under} it to get the segmented object. At farther distances, this procedure did not work as the plane was not flat anymore and thus RANSAC was unable to find the correct plane. Therefore, we manually segmented the object from the surroundings. Ultimately, we checked all individual point clouds for correct segmentation.
\end{itemize}

To evaluate the results, we needed to fit the segmented point clouds to a ground truth. We utilized the \ac{icp}~\cite{Besl92_ICP} algorithm for an initial transformation. The accuracy of \ac{icp} is limited when used only on partial point clouds of relatively small size (as in our case). Thus, for each captured and ground-truth point cloud we ran the algorithm with different convergence and inliers parameters and selected the best one as the one with the lowest distance between the estimated corresponding points. However, as this may still do not guarantee correct results at all times---especially at farther distances, where the objects are small and noisy. Thus, we further manually checked every estimated transformation and refined it to fit the ground truth properly. We used more than one frame for each object (for each camera at every depth), but we used only one transformation for the given setting. Otherwise, \ac{icp} could change the accuracy results, as it could mitigate the error by a transformation.

\subsection{Ground-truth Point Clouds}
For some metrics, ground-truth point clouds (shapes) are needed. For planar surfaces, we simply created a point cloud of a square with a side of 10\;cm. For the doll, there is no ground truth available. We utilized photogrammetry to create the model of the doll using Meshroom~\cite{alicevision2021}. The process consisted of taking pictures of the doll from various angles and view-points. Meshroom then performed feature matching using \ac{sift} from the unordered set of images. The features were then used to match pictures looking at the same areas of the scene and \ac{sfm} algorithm was used to create a 2D point cloud of the scene. Further, fronto-parallel planes were used to estimate depth for each of the pixels from \ac{sfm}. Finally, the resulting 3D point cloud was meshed using Delaunay tetrahedralization and textured using the input image. To add scale to the model, we put markers beside the doll with a known distance and we then scaled the model accordingly. We then manually refined the final 3D triangular mesh to remove unwanted holes and artifacts.

\begin{figure}[htb]
    \centering
    \includegraphics[width=1\columnwidth]{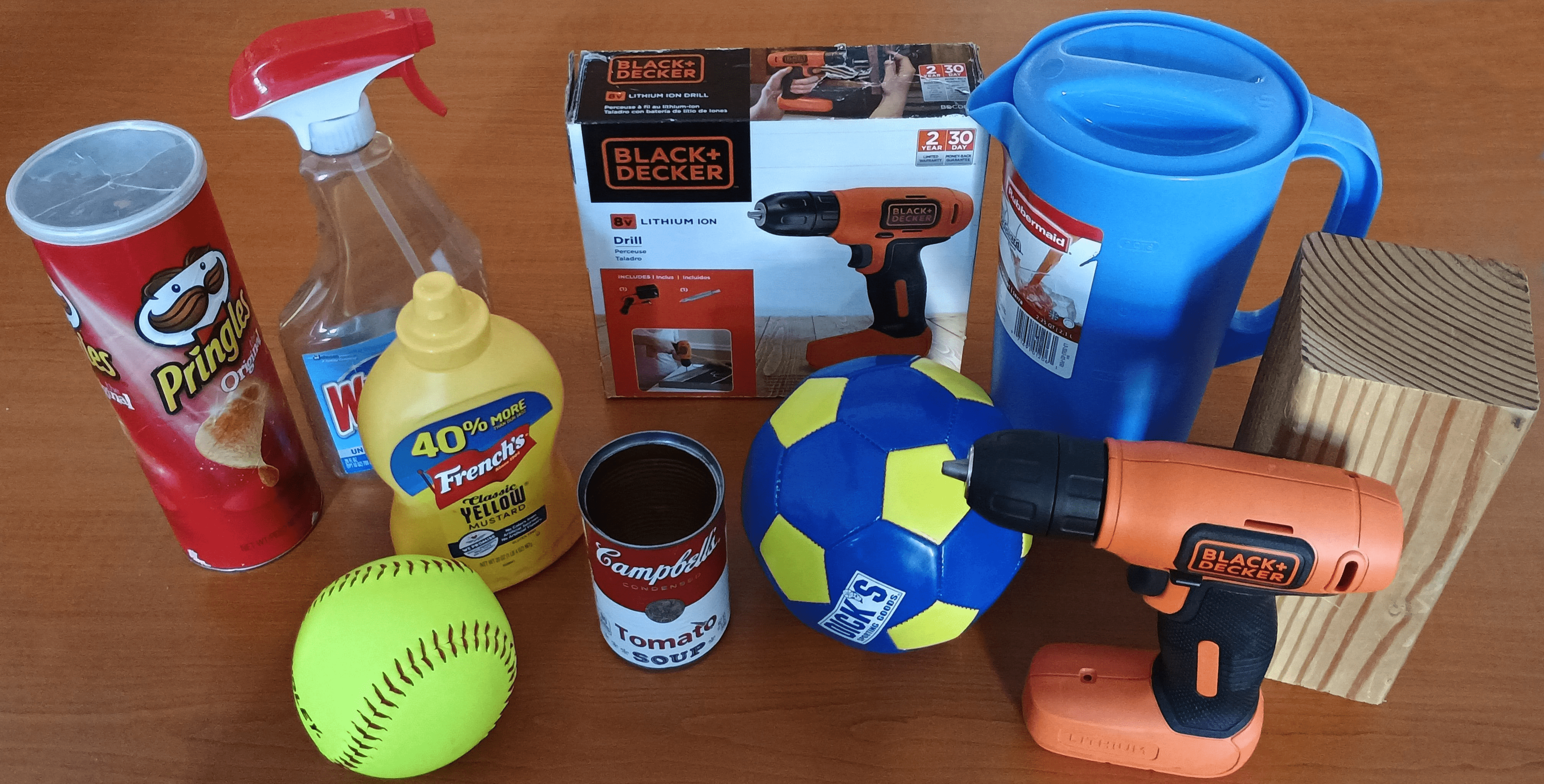}
    \caption{Objects from the YCB dataset used in the experiments.}
    \label{fig:objects}
\end{figure}

The YCB dataset was selected because it provides real-world items with corresponding models of the objects---see the used objects in \figref{fig:objects}. Thus, for objects from YCB we downloaded the corresponding model and sampled them into point clouds. For the transparent spray bottle, ground truth is not available from YCB, so we used photogrammetry as in the case of the plastic doll.

The ground truth point clouds for the YCB objects are complete point clouds from all sides. However, in our experiments, we obtained point clouds from one view only. To make the comparison fair, we manually cropped the ground-truth point clouds to match the viewpoint of the depth cameras. 

\subsection{Evaluation and Metrics}
We selected a set of metrics that compare different aspects of performance in all scenarios. Not all of them can be used in all scenarios. The definitions of the metrics are given below.

\subsubsection{Bias and Standard Deviation}
\label{sec:bias_precision}
This metric is also used in \cite{halmetschlager-funekEmpiricalEvaluationTen2019, heinemann2022MetrologicalApplicationrelatedComparison}. Bias describes the error between the mean estimated distance by the sensors and the ground-truth distance. It is defined as
\begin{equation}
\label{eq:bias}
    bias = |d_{gt} - \mu|,
\end{equation}
where $d_{gt}$ is the ground-truth distance and $\mu$ is the mean distance from the sensor defined as
\begin{equation}
    \label{eq:mu}
    \mu = \frac{1}{N\cdot M}\sum_{i=1}^{N}\sum_{j=1}^{M}d_{i, j},
\end{equation}
where $N$ is number of frames, $M$ is number of points in the given frame and $d_{i,j}$ is depth at point $j$ of frame $i$.

The standard deviation (SD) of depth measurements is defined as
\begin{equation}
\label{eq:precision}
    SD = \sqrt{\frac{1}{N\cdot M}\sum_{i=1}^{N}\sum_{j=1}^{M}(d_{i, j}-\mu)^2},
\end{equation}
where $\mu$ is from \eqref{eq:mu}.

This metric is used only for planar surface perception.

\subsubsection{Chamfer Distance}
\acf{cd} is a standard metric for estimating the distance of two point clouds. It is defined as the average distance of each point in one set to the closest point in the second set
and vice versa, i.e.,

\begin{align}
    d_{CD}(\mathbf{S}_1, \mathbf{S}_2) = \frac{1}{N}\sum_{\mathbf{x} \in \mathbf{S}_1}\min_{\mathbf{y} \in \mathbf{S}_2}\|\mathbf{x}-\mathbf{y}\|_2 + \\ \frac{1}{M}\sum_{\mathbf{y} \in \mathbf{S}_2}\min_{\mathbf{x} \in \mathbf{S}_1}\|{\mathbf{x}-\mathbf{y}}\|_2,
\end{align}
where $N, M$ are number of elements in $\mathbf{S}_1$ and $\mathbf{S}_2$, respectively. In this case, $\mathbf {S}_1$ and $\mathbf{S}_2$ are segmented point clouds from the camera and ground-truth point clouds, respectively.
This metric is used for plastic doll and YCB objects perception.

\subsubsection{Jaccard Similarity}
\acf{js} is a standard metric for estimating the geometric similarity of two shapes---with values from 0 to 1. It is defined as intersection over union of voxelized shapes, i.e., 
\begin{align}
    J(\mathbf{S}_1,\mathbf{S}_2) = \frac{|\mathbf{S}_1 \cap \mathbf{S}_2|}{|\mathbf{S}_1 \cup \mathbf{S}_2|},
\end{align}
 where $\mathbf{S}_1, \mathbf{S}_2$ are two sets. In our case, $\mathbf{S}_1$ and $\mathbf{S}_2$ are segmented and voxelized point clouds from the camera and ground-truth point clouds, respectively.

This metric is used for plastic doll and YCB objects perception. The higher the value, the better.

\subsubsection{F-score}
$F_1$ score is a harmonic mean of the precision and recall measures used in classification---with values from 0 to 1. We used the definition from NVIDIA Kaolin Library~\cite{KaolinLibrary}, where \ac{tp} is defined by the existence of two points in a given radius $r$ from each other. \ac{fn} are points in the camera point cloud that are further than $r$ from the closest point in ground truth and \ac{fp} are points from the ground truth that are further than $r$ from the camera point cloud. And then
\begin{equation}
    F_1 = \frac{2TP}{2TP + FP + FN}.
\end{equation}

This metric is used for YCB objects and plastic doll perception. The higher the value, the better.

\subsubsection{Angle between Normals}
This metric shows the angle between normals of two closest points in estimated and ground-truth point clouds. For each point in the camera point cloud, we find the closest point in the ground-truth point cloud and compute the angle as

\begin{equation}
    \theta = \arccos{\frac{n_1 \cdot n_2}{|n_1|\cdot|n_2|}},
\end{equation}
where $n_1, n_2$ are normals of two closest points.

\begin{figure}[htb]
    \centering
    \includegraphics[width=0.35\columnwidth]{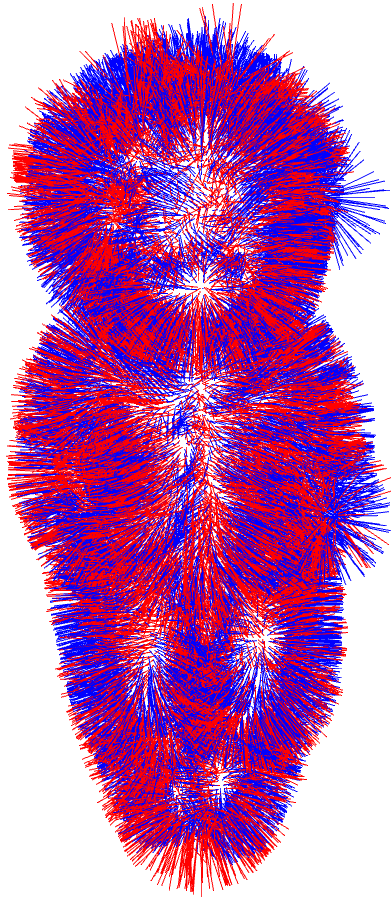}
    \caption{Normals of the ground-truth (blue) and captured (red) point cloud of the plastic doll.}
    \label{fig:normals_example}
\end{figure}

An example is shown in \figref{fig:normals_example}, where blue lines show normals of the ground-truth point cloud and red lines show normals of the captured point cloud. 

This metric is used for all scenarios.

\section{Experiments and Results}
\label{sec:experiments_and_results}
We tested the four cameras in three different scenarios: (i) planar surface perception, (ii) plastic doll perception, (iii) YCB objects perception. We measured each of these at several distances. For every object experiment, we used 30 successive frames, converted them to point clouds, and segmented the object out of the frame. In total, we compared the cameras with more than 3,000 point clouds from each camera, resulting in more than 12,000 total point clouds. All segmented point clouds, together with depth and RGB images, can be found at \url{https://rustlluk.github.io/rgbd-comparison}.

\subsection{Planar Surface Perception}
\begin{figure}[htb]
    \centering
    \includegraphics[width=\columnwidth,trim={0.5cm 0 2cm 0},clip]{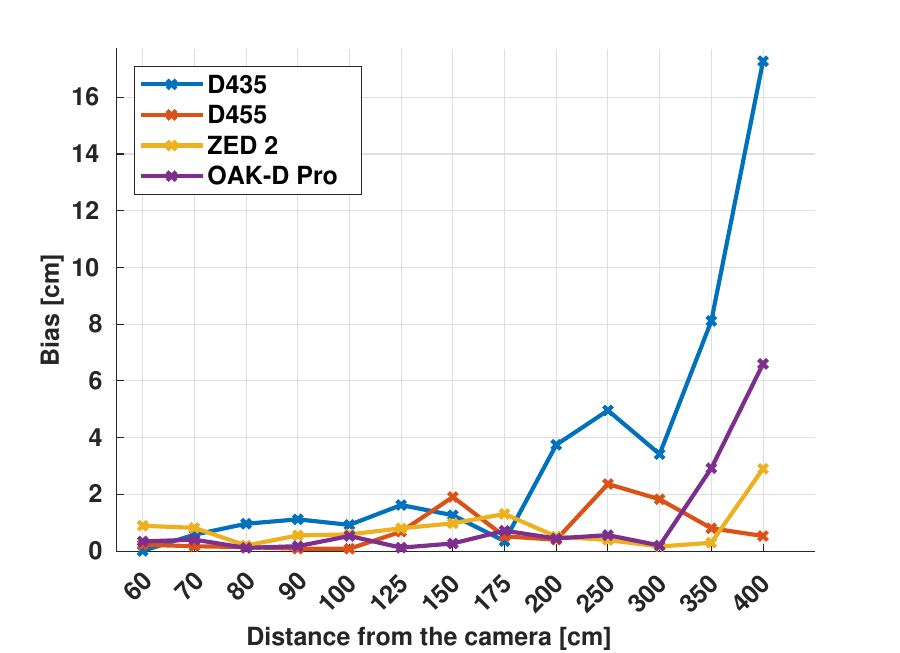}
    \includegraphics[width=\columnwidth,trim={0.5cm 0 2cm 0},clip]{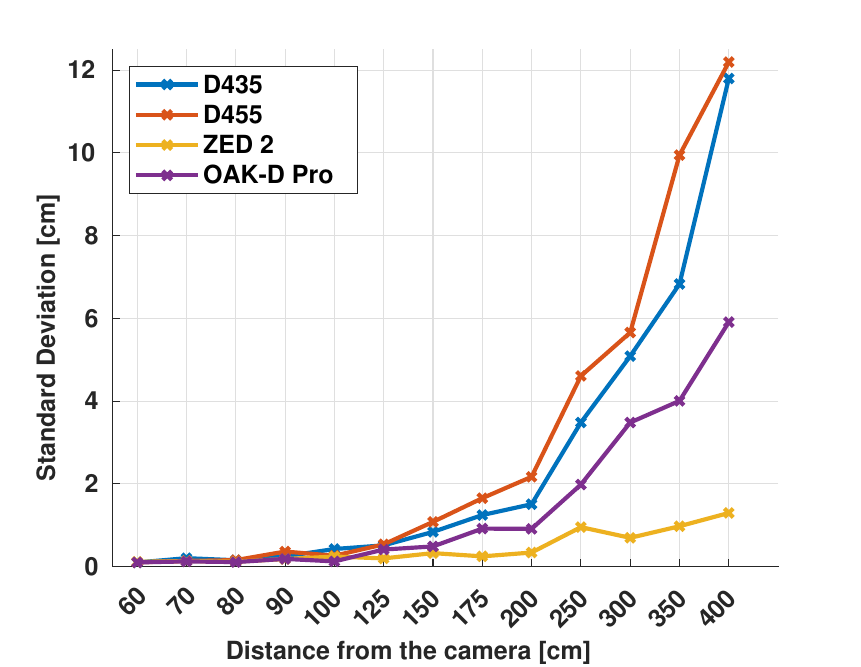}
    \caption{Planar surface perception -- bias and standard deviation. The values for each distance and camera are averaged from 30 frames. (Top) Bias in distance estimation to the plane (0 bias is correct). (Bottom) -- Standard deviation of the estimates.}
    \label{fig:bias_precision}
\end{figure}

We recorded samples at 60, 70, 80, 90, 100, 125, 150, 175, 200, 250, 300, 350, and 400\;cm from the camera. \figref{fig:bias_precision} shows the bias and standard deviation (defined in \secref{sec:bias_precision}) for $10$ $\times$ 10\;cm squares extracted from 30 consecutive frames taken by the cameras. For all cameras,  the bias (error of the mean distance) is small ($<$2\;cm) up to 175\;cm. Then, the bias of D435 increases significantly, mainly beyond 300\;cm, corresponding to the max. range of the D435. The OAK-D Pro bias starts to deviate after 300\;cm. For ZED 2 and D455, the bias is $<$3\;cm even at the farthest distance. 

For the standard deviation of distance from 30 measurements, all cameras behave similarly up to 100\;cm with error $<$0.5\;cm. Then, the RealSense devices (D435, D455) start to lose precision; this is also the case for the OAK-D Pro but to a lesser extent. In case of ZED 2, the standard deviation is $<$1.5\;cm for all distances. We can also compare with the accuracy provided by the manufacturers. The comparison is shown in \tabref{tab:factory_vs_real}. It is not always known what the term \enquote{accuracy} means for each of the cameras, but we assume that it will be comparable to either bias or standard deviation. We can see that the real errors are lower than the ones reported by the manufacturers, except for the standard deviation of D455.

\begin{table}[htb]
\caption{Planar surface perception -- comparison with factory accuracy.}
\label{tab:factory_vs_real}
\centering
\resizebox{\columnwidth}{!}{%
\begin{tabular}{c|c|c|c|}
\cline{2-4}
 & \textbf{Factory accuracy} & \textbf{Standard deviation} & \textbf{Bias} \\ \hline
\multicolumn{1}{|c|}{\textbf{D435}} & $<$2\% at 2\;m & $0.75$\% at 2\;m & $1.87$\% at 2\;m \\ \hline
\multicolumn{1}{|c|}{\textbf{D455}} & $<$2\% at 4\;m & $3.05$\% at 4\;m & $0.13$\% at 4\;m \\ \hline
\multicolumn{1}{|c|}{\textbf{ZED}} & $<$0.8\% at 2\;m & $0.17$\% at 2\;m & $0.22$\% at 2\;m \\ \hline
\multicolumn{1}{|c|}{\textbf{OAK-D Pro}} & $<$2\% at 4\;m & $1.47$\% at 4\;m & $1.65$\% at 4\;m \\ \hline
\end{tabular}
}
\end{table}

\begin{figure}[htb]
    \centering
    \includegraphics[width=1\columnwidth]{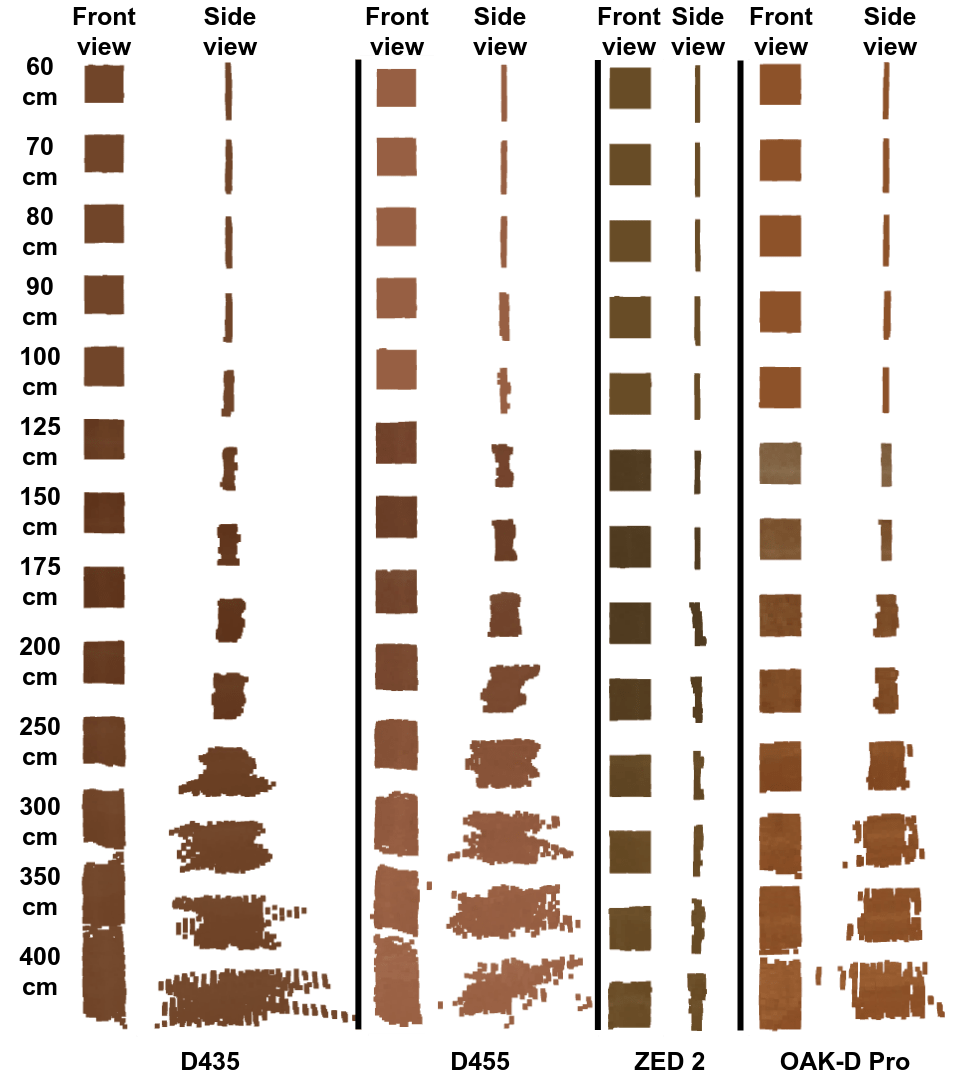}
    \caption{Planar surface perception -- point clouds. Front (left columns) and side (right columns) views on point clouds of the extracted planes (concatenated over all 30 frames) for all distances.}
    \label{fig:planes}
\end{figure}

The standard deviation can also be seen in the right columns of \figref{fig:planes}, where it is represented as the scatter of the points on the sides. It is well visible that, except for ZED 2, the width of the planes increases with increasing distance, corresponding to an increase in standard deviation in \figref{fig:bias_precision}. 

\begin{figure}[ht]
    \centering
    \includegraphics[width=\columnwidth,trim={0.65cm 0 1.5cm 0},clip]{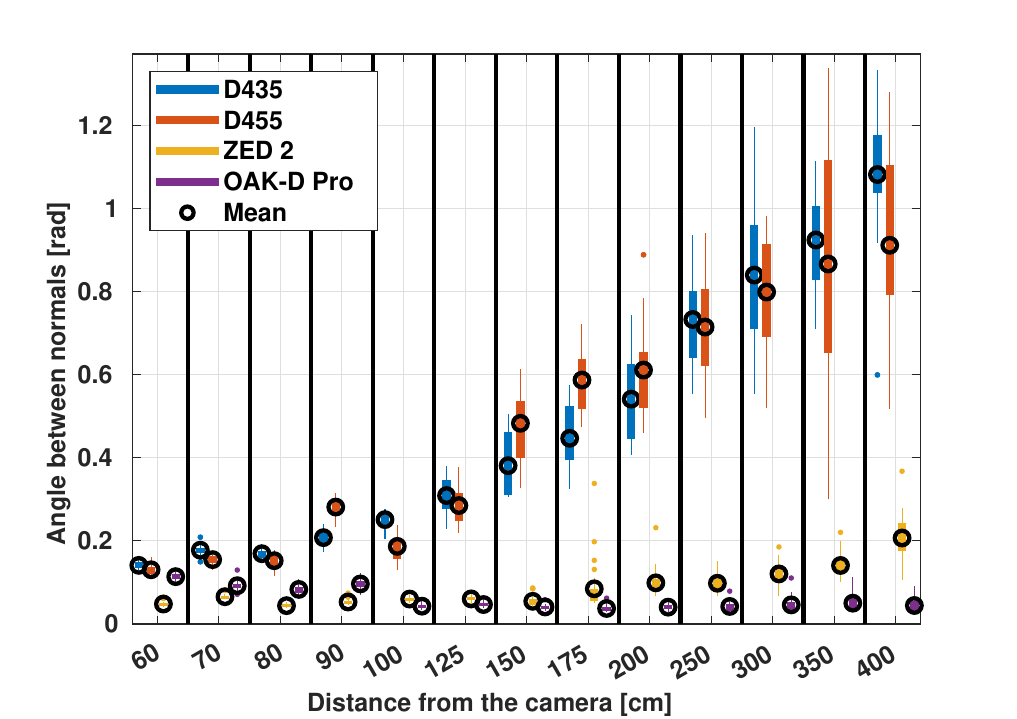}
    \caption{Planar surface perception -- angles between the closest normals. Bars show values from 25th to 75th percentile; whiskers represent non-outlier extreme points and dots represent outliers. Computed over 30 frames for each distance and camera. The lower the value, the better.}
    \label{fig:planes_normals}
\end{figure}

Another metric usable for plane perception is the angle between the closest normals in camera and ground-truth point clouds. In case of planes, this metric is basically a metric of flatness. The results can be seen in \figref{fig:planes_normals}. Up to 80\;cm all cameras stay under 0.2\;rad (\textasciitilde10$^\circ$). However, the RealSense devices have twice the error even at the lower distances. At higher distances, the error of the RealSense devices increases to around 1\;rad (\textasciitilde\ang{60}). ZED 2 stays $<$0.2\;rad and OAK-D Pro $<$0.1\;rad for all distances. Farther than 100 cm from the camera, the RealSense devices (D435, D455) have more variability over different frames, while OAK-D Pro and ZED 2 show only small variability. The planes extracted from D435 and D455 are more uneven with increasing distance. The planes extracted from ZED 2 and OAK-D Pro are equally flat for all distances. In fact, OAK-D Pro always returns points in almost \enquote{flat layers}, and thus the error from a flat surface is low.

\textbf{Planar surface perception -- summary.} All the results \figref{fig:bias_precision}, \figref{fig:planes}, \figref{fig:planes_normals} paint a coherent picture. For distances above one meter, D435 and D455 start losing both accuracy (positive bias -- distance overestimation) and precision (greater variability). ZED 2 has very good performance even for the maximum tested distance (400 cm). Planar surface perception is an important task for mobile navigation, mainly in indoor environments. Based on the results, the best choice in this scenario would be D435, OAK-D Pro, or ZED 2. The choice between these is then mainly related to the fact whether you can put a dedicated GPU on the robot or not.

\subsection{Plastic Doll Perception}
\label{sec:doll}

\begin{figure}[htb]
    \centering
    \includegraphics[width=1\columnwidth]{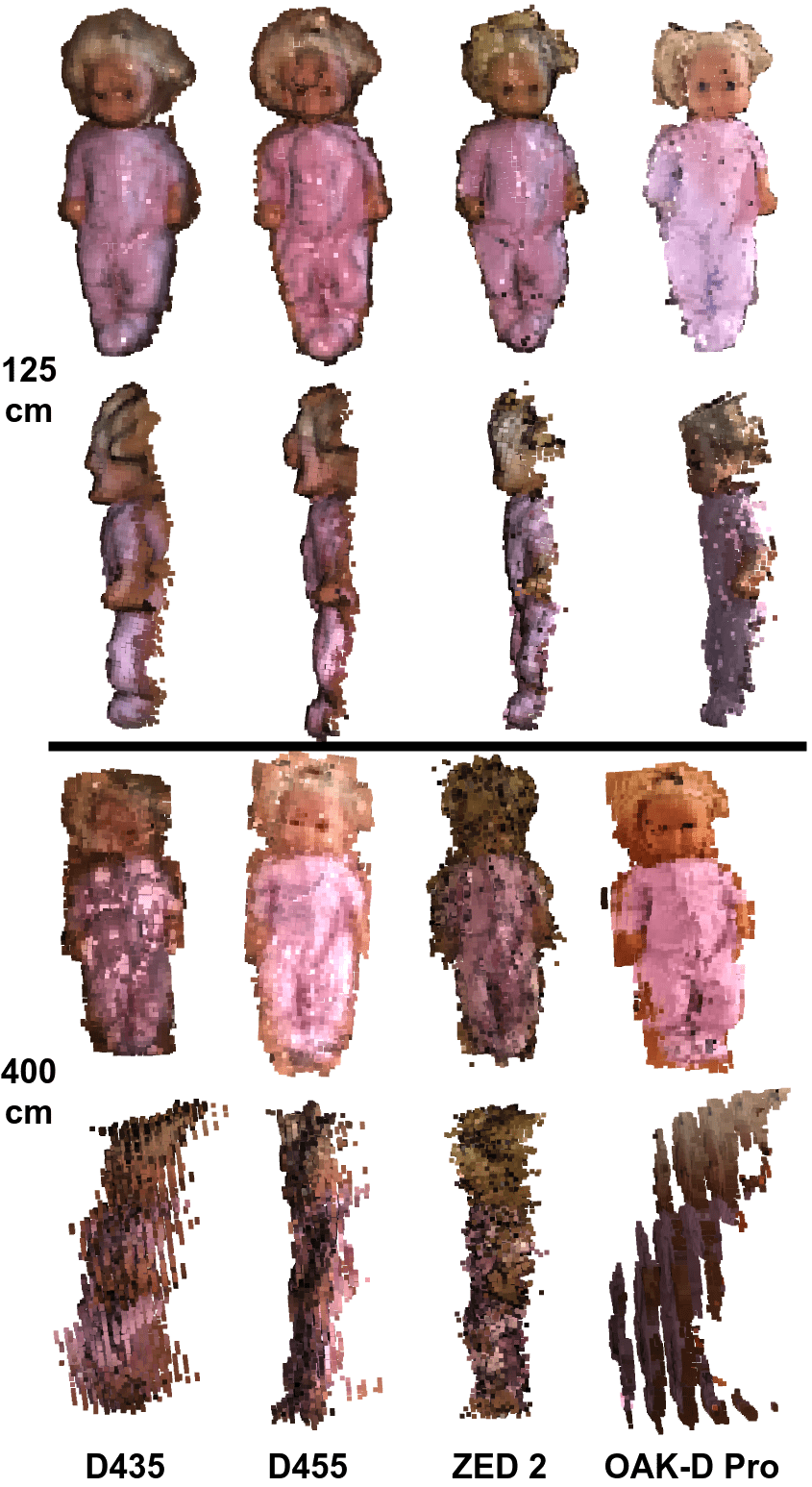}

    \caption{Plastic doll perception -- point clouds. At 125\;cm (upper block) and 400\;cm (lower block) distance. Front and side views on the point clouds (concatenated over all 30 frames).}
    \label{fig:dolls}
\end{figure}

The second set of experiments involved the perception of a plastic doll shown in \figref{fig:setup}. We recorded the samples at 90, 100, 125, 150, 175, 200, 250, 300, 350, and 400\;cm from the camera. Point clouds taken from 125\;cm and 400\;cm can be seen in \figref{fig:dolls}. We can see that at the closer distance, the point clouds look visually fine from both front and side views. The most visible flaws are at the forehead, which is a challenging part for RGB-D cameras as it is featureless and shiny. On the other hand, in the farther distance the point clouds are much noisier. The noise is visible especially from the side view, where the uneven depth estimation is visible for all cameras. However, it is interesting how different cameras handle the depth differently. RealSense devices divide the depth into layers with small differences between them and include some noisy points. OAK-D Pro divides the distances into layers with higher distances without visible noise. ZED 2 has the lowest depth difference among individual frames, but has the most noisy points. 

\begin{figure}[htb]
    \centering
    \includegraphics[width=1\columnwidth,trim={0.75cm 0 1.5cm 0},clip]{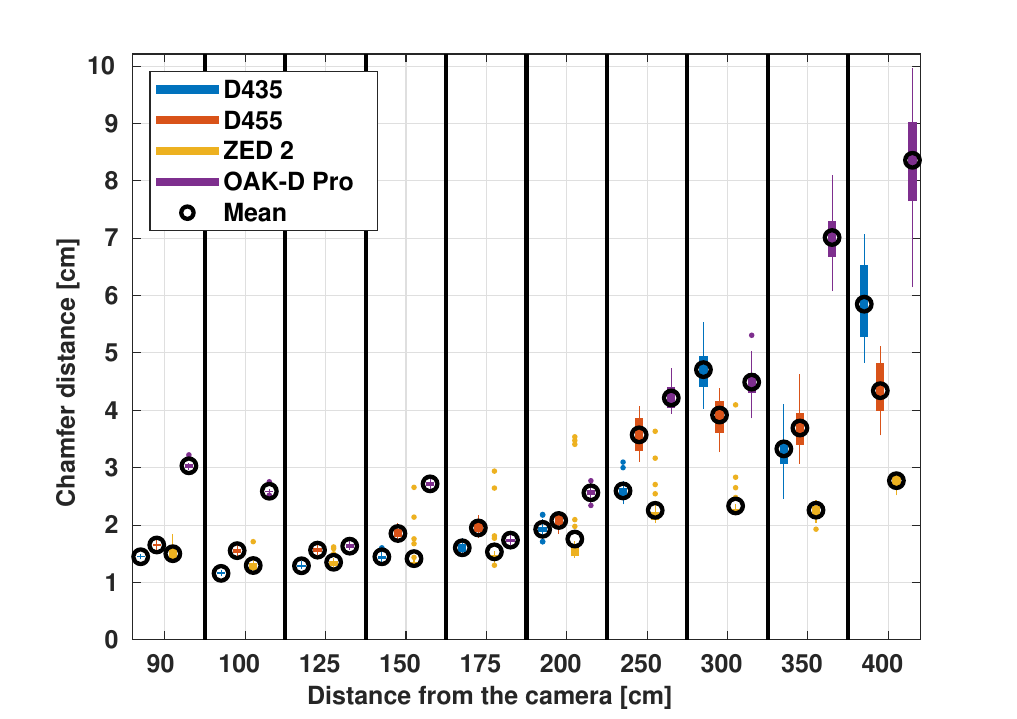}

    \caption{Plastic doll perception -- \acf{cd}. Bars show values from 25th to 75th percentile; whiskers represent non-outlier extreme points and dots represent outliers. Computed over 30 frames for each distance and camera. The lower the value, the better.}
    \label{fig:dolls_cham}
\end{figure}

The first empirical metric we can look at is the \ac{cd} shown in \figref{fig:dolls_cham}. Note that at 90 and 100\;cm, the OAK-D Pro camera did not see the entire doll, thus the comparison at these distances is not fair. Otherwise, until 175\;cm all cameras stay under 2\;cm error, with ZED 2 having the best performance, followed by D435. Interestingly, D455 behaves worse than D435. We assume that it is caused by a higher baseline of D455, which helps for flat surfaces but has a negative effect for more complicated surfaces. The error rises steeply after that for all cameras except ZED 2. At 400\;cm ZED 2 still keeps the \ac{cd} under 3\;cm, whereas OAK-D Pro has more than 8\;cm error. It is caused mainly by the layering of depth---visible in \figref{fig:dolls}. At the final distance, D455 is finally better than D435, but, as has been said, this distance is above the ideal range of D435. The variance between individual frames is gradually increasing for all cameras except for ZED 2.

\begin{figure}[htb]
    \centering
    \includegraphics[width=1\columnwidth,trim={0.5cm 0 1.5cm 0},clip]{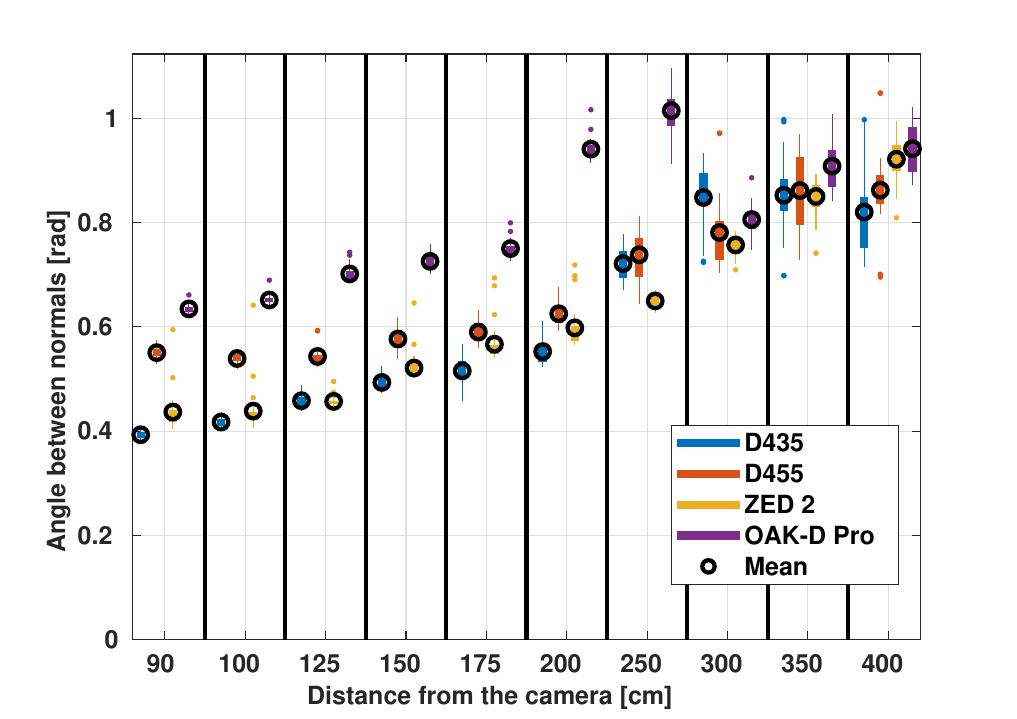}

    \caption{Plastic doll perception -- angles between the closest normals. Bars show values from 25th to 75th percentile, whiskers represent non-outlier extreme points, and dots represent outliers. Computed over 30 frames for each distance and camera. The lower the value, the better.}
    \label{fig:dolls_normals}
\end{figure}

\begin{figure}[htb]
    \centering
    \includegraphics[width=1\columnwidth,trim={1cm 0 1.5cm 0},clip]{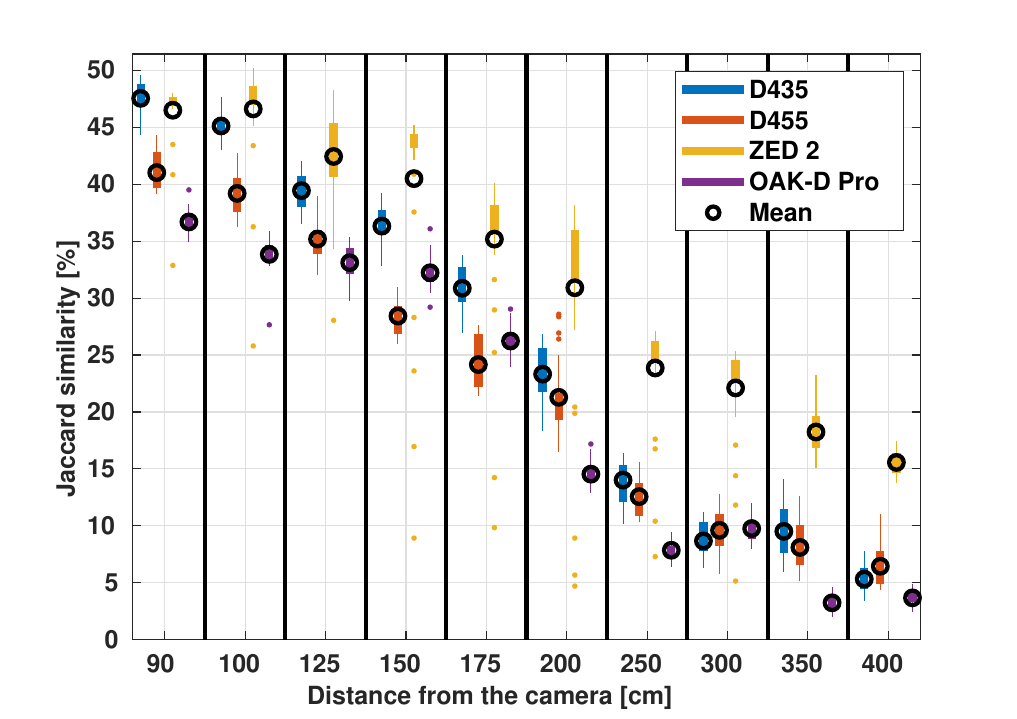}

    \caption{Plastic doll perception -- \acf{js}. Bars show values from 25th to 75th percentile, whiskers represent non-outlier extreme points and dots represent outliers. Computed over 30 frames for each distance and camera. The higher the value, the better.}
    \label{fig:dolls_jac}
\end{figure}

\begin{figure}[htb]
    \centering
    \includegraphics[width=1\columnwidth,trim={0.5cm 0 1.5cm 0},clip]{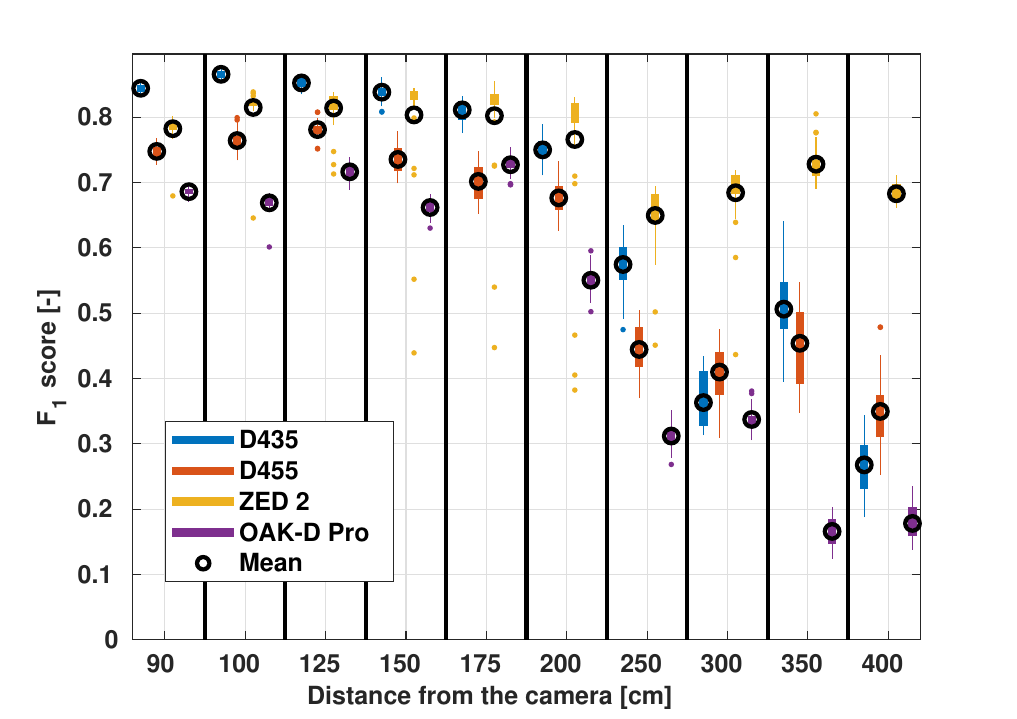}

    \caption{Plastic doll perception -- $F_1$ score. Bars show values from 25th to 75th percentile, whiskers represent non-outlier extreme points and dots represent outliers. Computed over 30 frames for each distance and camera. The higher the value, the better.}
    \label{fig:dolls_f}
\end{figure}

The other metric is the angle between normals. Here, it is not a measure of flatness as in the case of planar surfaces, but rather a measure of shape estimation. The results can be seen in \figref{fig:dolls_normals}. An important thing to notice is the high error of the OAK-D Pro camera at all depths. As has been said, OAK-D Pro creates layers of points at different depths (the depth differences between layers are lower for a lower object distance from the camera) instead of a scatter distribution, as is for other cameras. That results in facing of the normals more in the direction of the camera plane. Among the other cameras, ZED 2 and D435 are on par for every distance, and D455 is worse up to 150\;cm (again, we attribute this to the wider baseline), and then all cameras perform almost the same. Overall, the angles are higher than for the planar surfaces in lower distances---around 0.2\;rad vs. 0.4\;rad at the closest distance. However, it is interesting to look at the higher distances. RealSense devices achieve the same error as for the planes---about 1\;rad. Both ZED 2 and OAK-D Pro are on par with Intel devices, but in the case of planes the error for the highest distance was under 0.2\;rad for ZED 2 and 0.1\;rad for OAK-D Pro. That is a huge difference that shows how the devices sense simple (planar) and more complex environments differently.

\begin{figure*}[htb]
    \centering
    \includegraphics[width=\textwidth]{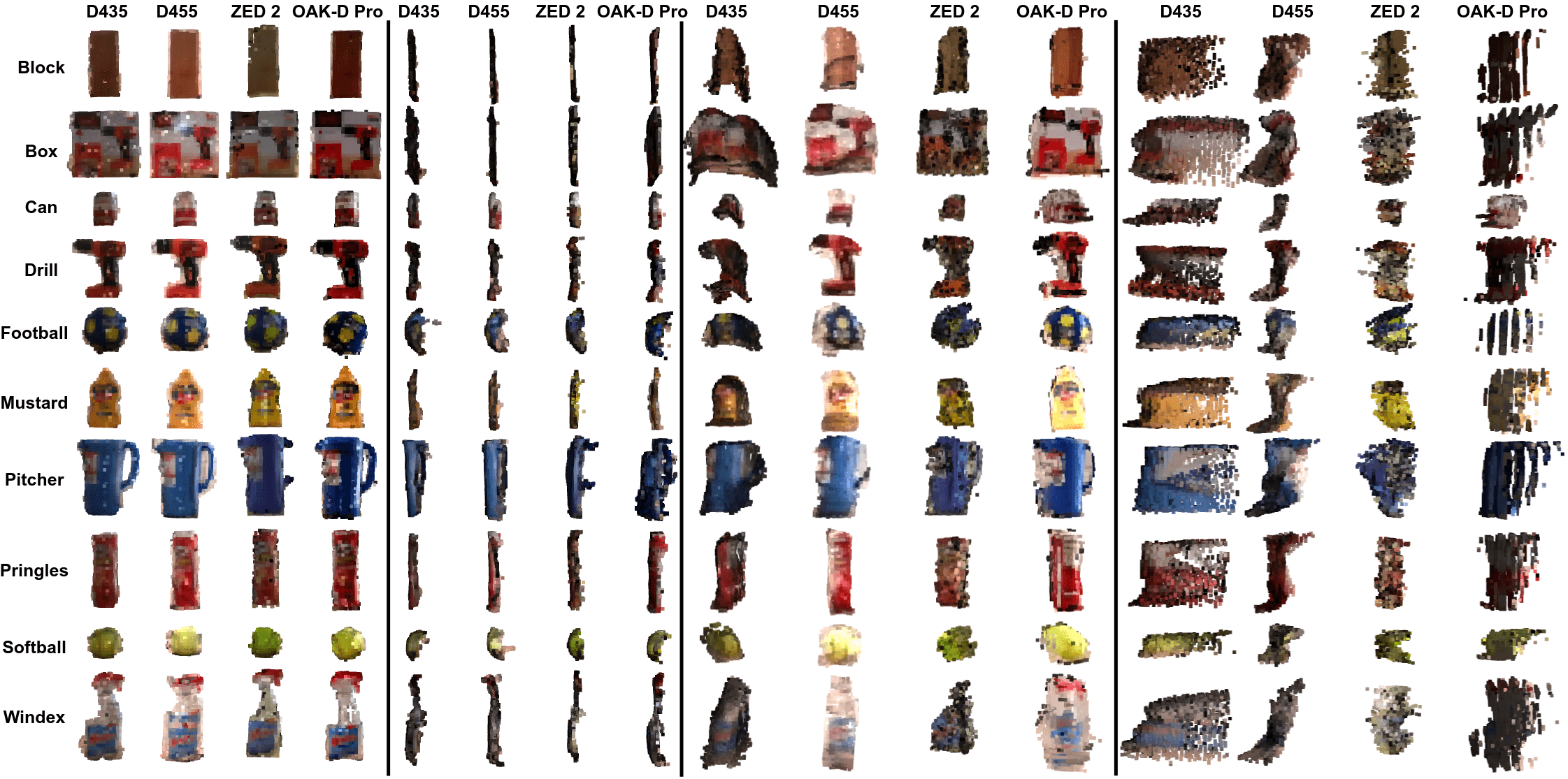}
    
    \begin{minipage}[t]{0.49\textwidth}
        \centering
        (a) 60\;cm 
    \end{minipage}
    \begin{minipage}[t]{0.49\textwidth}
        \hspace{-2.5em}
        \centering
        (b) 300\;cm
    \end{minipage}
    
    \caption{YCB objects perception -- point clouds. Front (left blocks) and side (right blocks) views for 60\;cm (a) and 300\;cm (b).}
    \label{fig:ycb}
\end{figure*}

For the final two metrics---\acf{js} and $F_1$ score---the higher the value, the better. Both of these metrics evaluate the similarity of the shapes. In case of \ac{js}, the output is \enquote{geometric closeness} of two shapes. For $F_1$, it is the probability that a point in a given point cloud should be where it is when considering the ground truth. For D435, D455, and OAK-D Pro, the results are similar, with D435 being the best among these. In case of \ac{js}, the similarity is decreasing with a steeper drop from 250\;cm further. $F_1$ similarity also demonstrates the same drop. For ZED 2, \ac{js} is getting gradually lower without any significant drop, and $F_1$ reveals only a slight drop (around 0.1) at higher distances. The reason is the amount of noise. As can be seen in the bottom block of \figref{fig:dolls}, the ZED 2 point clouds have the lowest variance in depth estimates between frames, and the noise is more concentrated around the true values, compared to other cameras that produce more scattered results.

\textbf{Plastic doll perception -- summary.}
All the cameras performs similarly well up to 150\;cm in all metrics. The best overall cameras is the ZED 2, that achieves \ac{cd} error $<3\;cm$ even at the farthest distance (400\;cm). The worst camera in this task is the OAK-D Pro. Both Intel devices perform similarly and up to 200\;cm they are comparable to the ZED 2 camera. This scenario simulates mainly the task of human pose and shape estimation. If humans are perceived from a close distance, typical for human-robot interaction scenarios, it is better to choose either D435 or ZED 2. ZED 2 directly offers keypoint detection algorithms.

\subsection{YCB Objects Perception}

For this setting, we recorded at 60, 80, 100, 125, 150, 200, 250, and 300\;cm from the cameras, where we used 10 different objects---see \figref{fig:objects} and \figref{fig:ycb}. The objects were selected to represent various shapes, sizes, and materials.

\figref{fig:ycb} includes front and side views of point clouds captured by the cameras at 60\;cm and 300\;cm. We can see that for the small distance, most of the objects are visually fine. Some flaws can be seen for \textit{can} and \textit{windex bottle}. The main difference between the cameras is for the \textit{pitcher}, where the best results come from the D435 and OAK-D Pro. D455 was unable to detect the hole in the handle, and ZED 2 missed half of the handle. The side views are worse again for the \textit{pitcher} and then for the \textit{windex bottle}. The bottle is transparent (see \figref{fig:objects}), which is an adversarial surface for RGB-D cameras. At 300\;cm, the best front views are generally distorted and noisy and the best are probably for D455 and ZED 2. The more interesting is the side view, which reveals a higher standard deviation in the depth estimation from D435 and OAK-D Pro. Point clouds for D455 and ZED 2 have less scatter in depth estimation, but ZED 2 has noisy points and D455 creates surface coplanar with the table at the bottom parts of the objects.

\begin{figure}[htb]
    \centering
    \includegraphics[width=1\columnwidth,trim={1cm 0 1.5cm 0},clip]{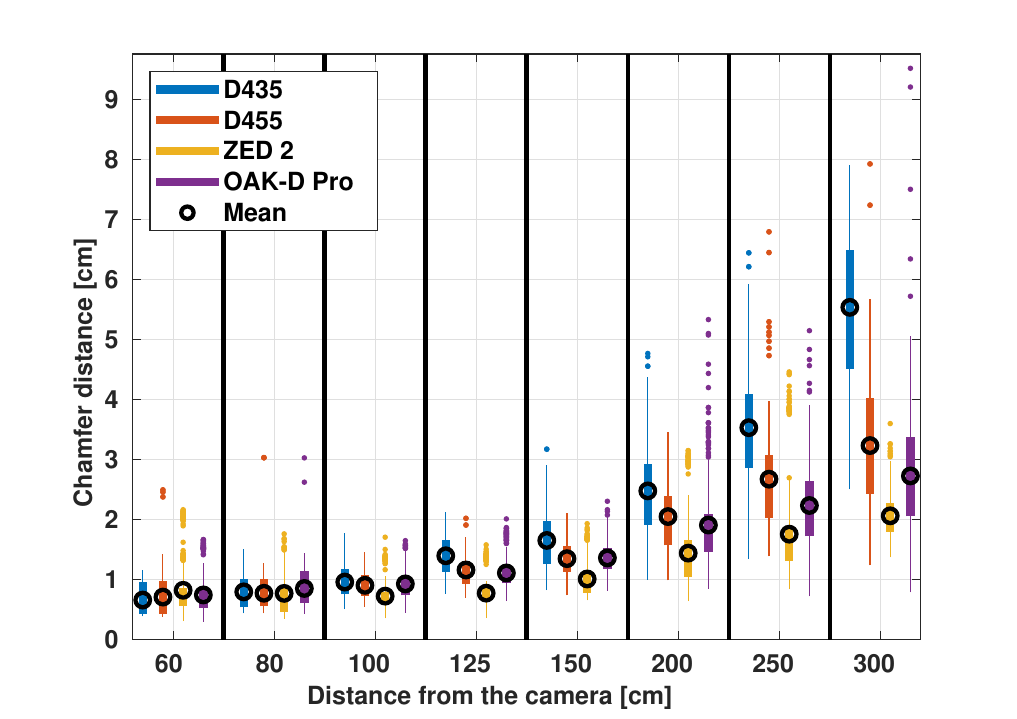}

    \caption{YCB objects perception -- \acf{cd}. Bars show values from 25th to 75th percentile, whiskers represent non-outlier extreme points and dots represent outliers. Computed over 30 frames of 10 objects for each distance and camera. the lower the value, the better.}
    \label{fig:ycb_cham}
\end{figure}

\acf{cd} for the YCB objects can be seen in \figref{fig:ycb_cham}. We can see that the trend is the same as in the case in the plastic doll perception. Up to 150\;cm all the cameras perform almost the same with \ac{cd} around 1\;cm. After that, the error for RealSense devices increases up to about 5.5\;cm for D435. The best is ZED 2 with about 2\;cm even at 300\;cm. The difference from the plastic doll perception is that OAK-D Pro is now the second best camera in terms of mean \ac{cd}. We assume that the reason is that some of the objects selected are flat, and thus OAK-D Pro performs well on these. But, as seen from high variance and outlier points, it performs the worst on some objects. 

\begin{figure}[htb]
    \centering
    \includegraphics[width=1\columnwidth,trim={0.75cm 0 1.5cm 0},clip]{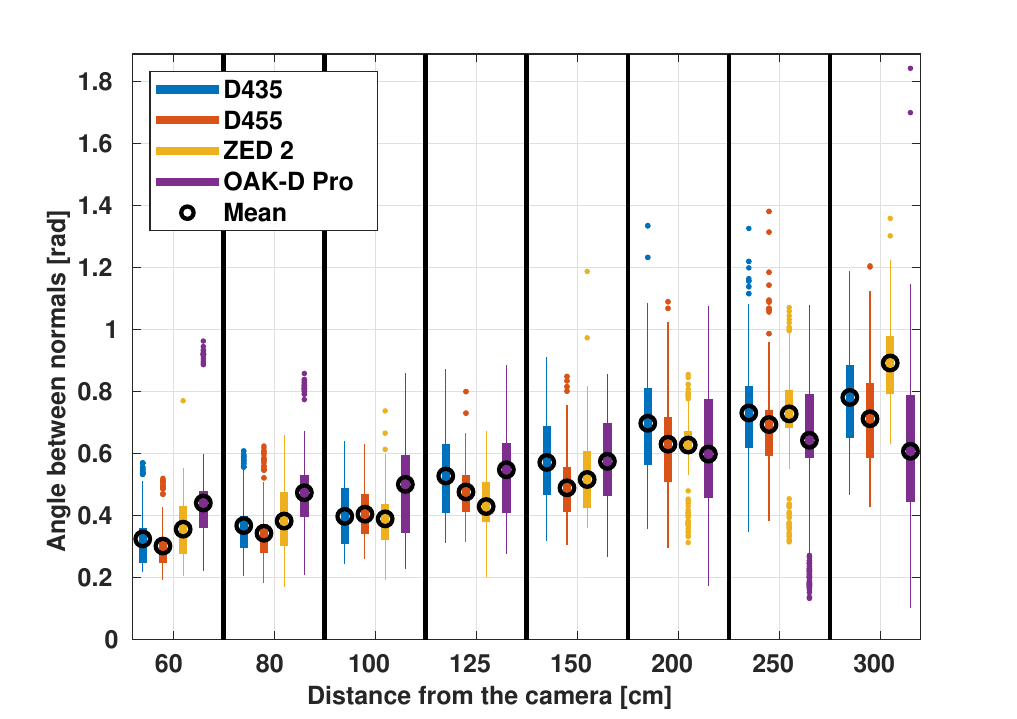}

    \caption{YCB objects perception -- angles between the closest normals. Bars show values from 25th to 75th percentile, whiskers represent non-outlier extreme points and dots represent outliers. Computed over 30 frames of 10 objects for each distance and camera. The lower the value, the better.}
    \label{fig:ycb_normals}
\end{figure}

Angles between the closest normals for YCB objects are shown in \figref{fig:ycb_normals}. The results are similar to the angles for the plastic doll perception (see \figref{fig:dolls_normals}), with the difference for OAK-D Pro. In case of plastic doll perception, there was a big difference between OAK-D Pro and the other cameras. Here, all cameras have similar results. This is again caused by the fact that some objects selected are flat, and therefore OAK-D Pro can estimate these well. However, the outliers (mainly at 60\;cm) shows that it has problems with some objects.

\begin{figure}[htb]
    \centering
    \includegraphics[width=1\columnwidth,trim={1cm 0 1.5cm 0},clip]{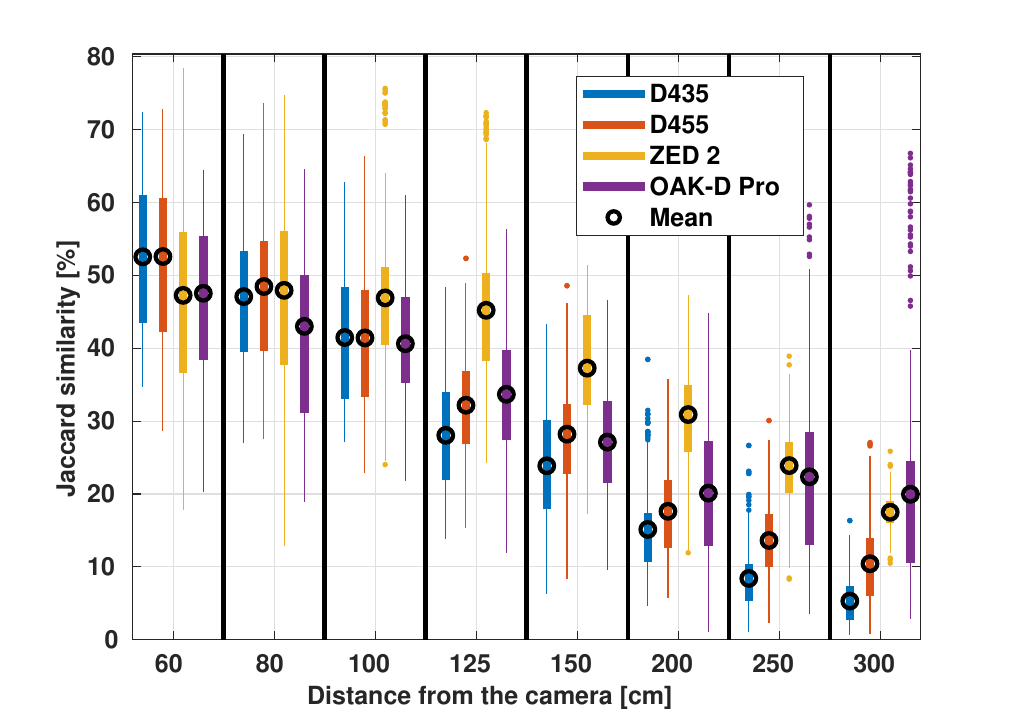}

    \caption{YCB objects perception -- \acf{js}. Bars show values from 25th to 75th percentile, whiskers represent non-outlier extreme points and dots represent outliers. Computed over 30 frames of 10 objects for each distance and camera. The lower the value, the better.}
    \label{fig:ycb_jac}
\end{figure}

\begin{figure}[htb]
    \centering
    \includegraphics[width=1\columnwidth,trim={0.5cm 0 1.5cm 0},clip]{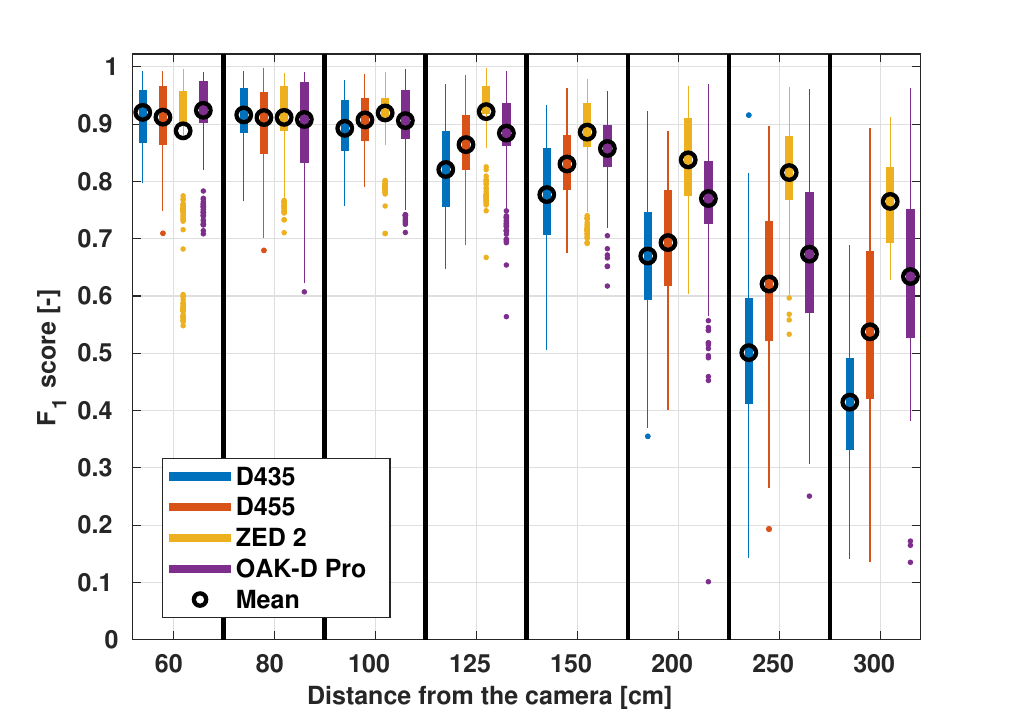}
    \caption{YCB objects perception -- $F_1$ score. Bars show values from 25th to 75th percentile, whiskers represent non-outlier extreme points and dots represent outliers. Computed over 30 frames of 10 objects for each distance and camera. The lower the value, the better.}
    \label{fig:ycb_f}
\end{figure}

\acf{js} for this scenario is shown in \figref{fig:ycb_jac}. We can see that at lower distances, the RealSense devices perform the best and ZED 2 is getting better with increasing distance. At higher distances, the mean is similar for both ZED 2 and OAK-D Pro, but based on the variance and outliers, ZED 2 performs comparably on all objects. OAK-D Pro, on the other hand, has many outliers---objects for which the estimation is good, and thus the mean \ac{js} is higher. Analogous results are visible for the $F_1$ score in \figref{fig:ycb_f}. At lower distances, all cameras achieve similar scores, but gradually ZED 2 is the best with the highest $F_1$ score and lower variance between objects.

\begin{figure*}[htb]
    \centering
    \begin{minipage}[t]{\columnwidth}
        \centering
        \includegraphics[width=\textwidth,trim={1.5cm 0 1.5cm 0},clip]{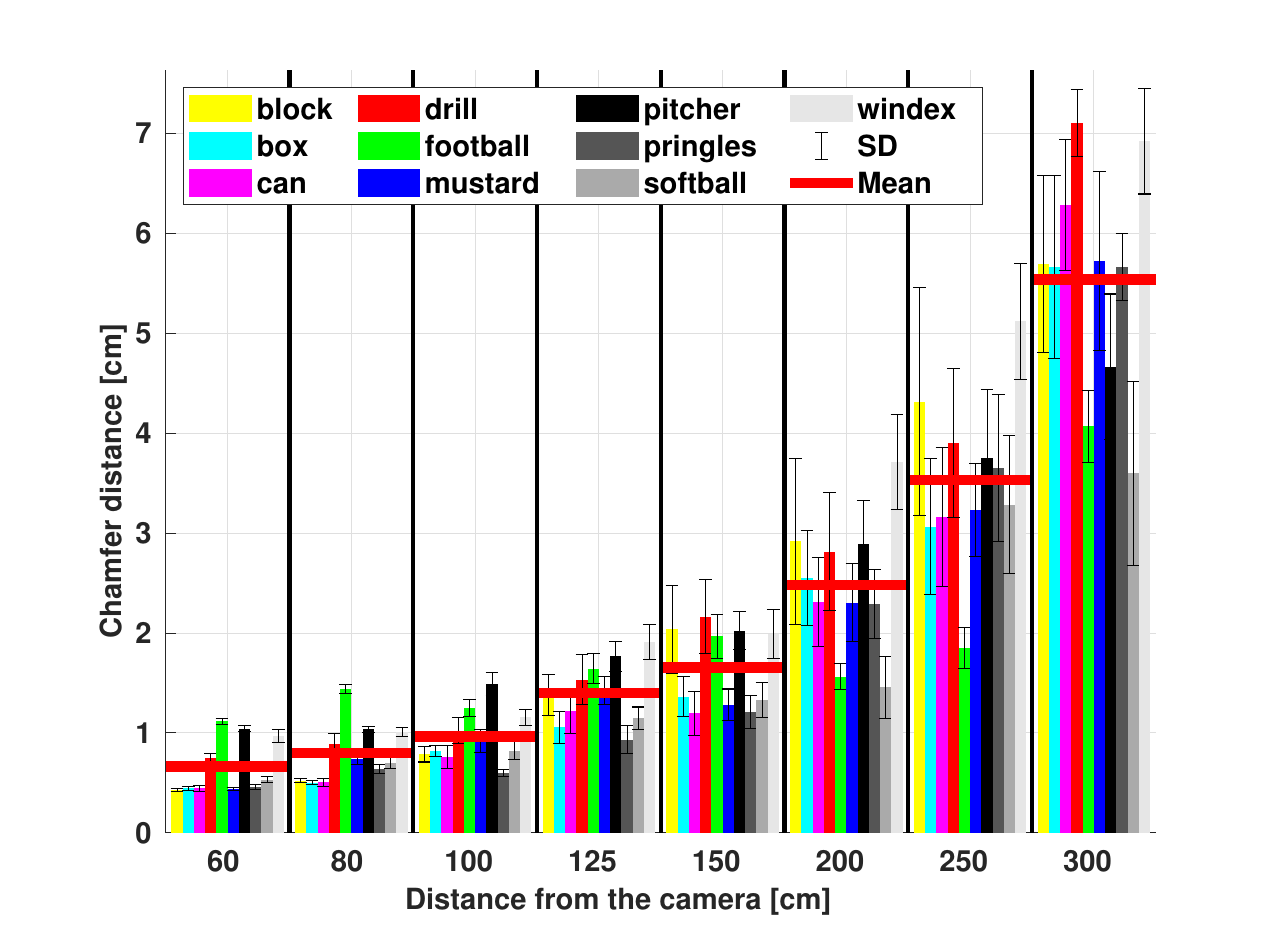}
        (a) D435.
    \end{minipage}
    \begin{minipage}[t]{\columnwidth}
        \centering
        \includegraphics[width=\textwidth,trim={1.5cm 0 1.5cm 0},clip]{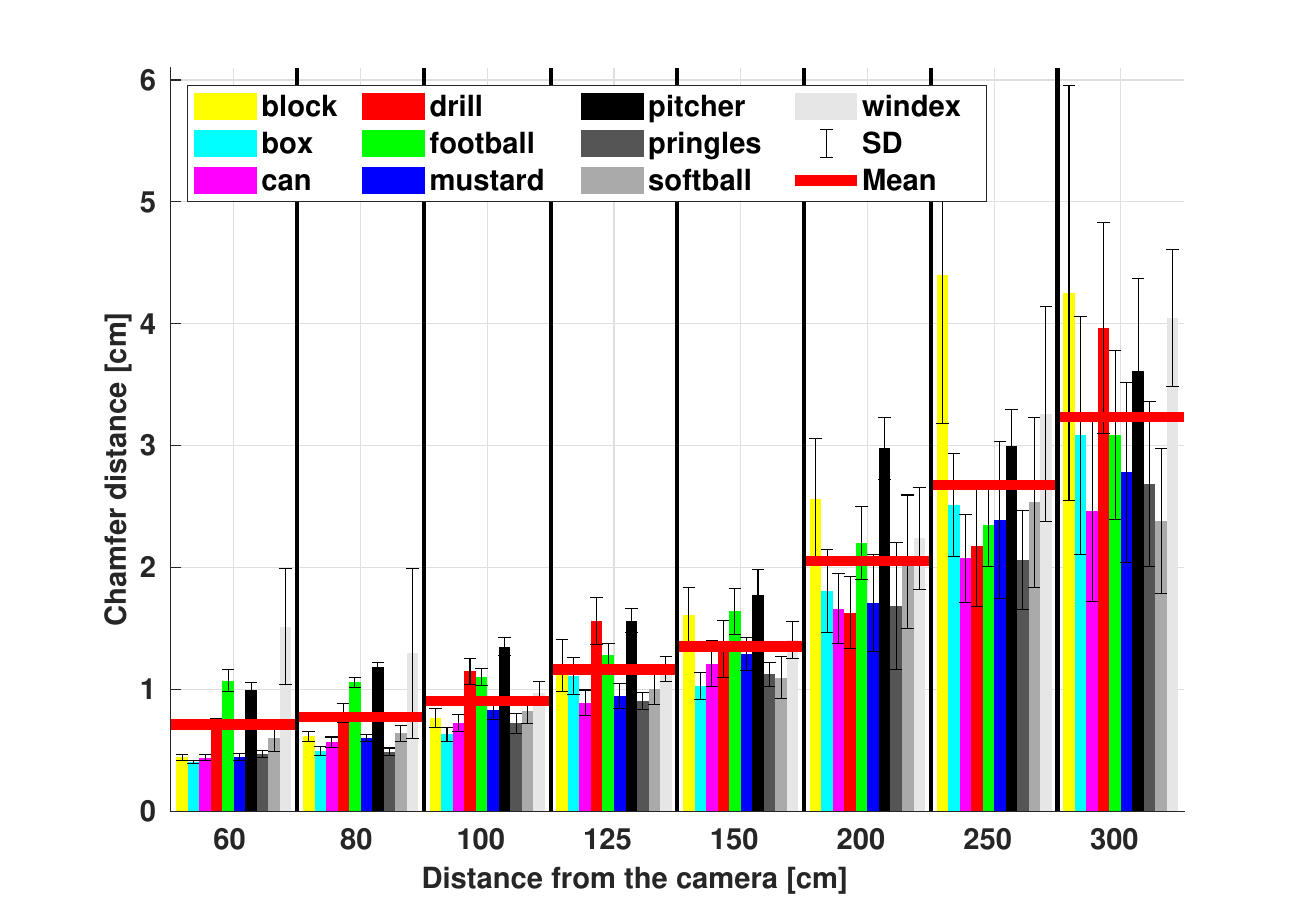}
        (b) D455.
    \end{minipage}
    \begin{minipage}[t]{\columnwidth}
        \centering
        \includegraphics[width=\textwidth,trim={1.2cm 0 1.5cm 0},clip]{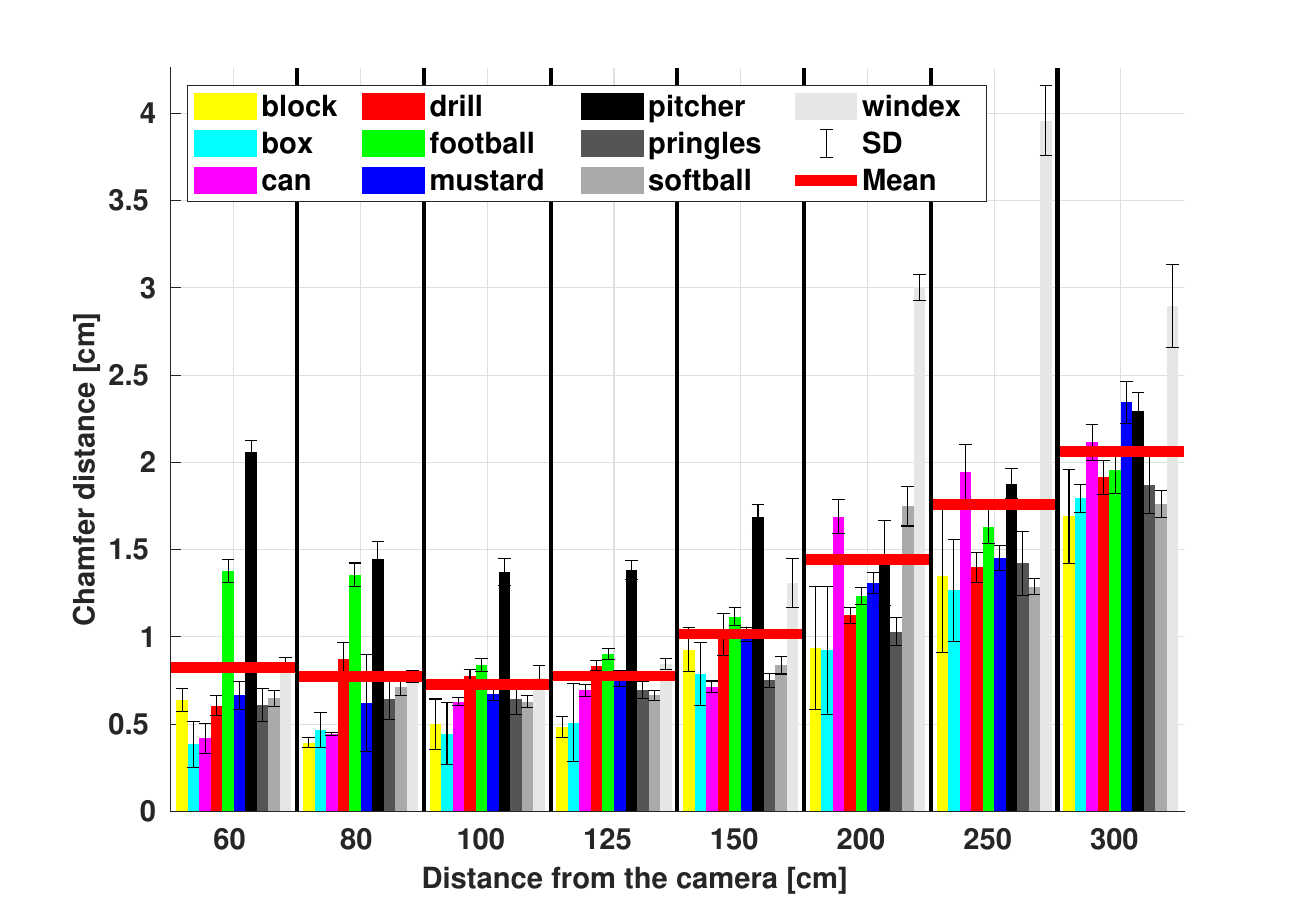}
        (c) ZED 2.
    \end{minipage}
    \begin{minipage}[t]{\columnwidth}
        \centering
        \includegraphics[width=\textwidth,trim={1.25cm 0 1.5cm 0},clip]{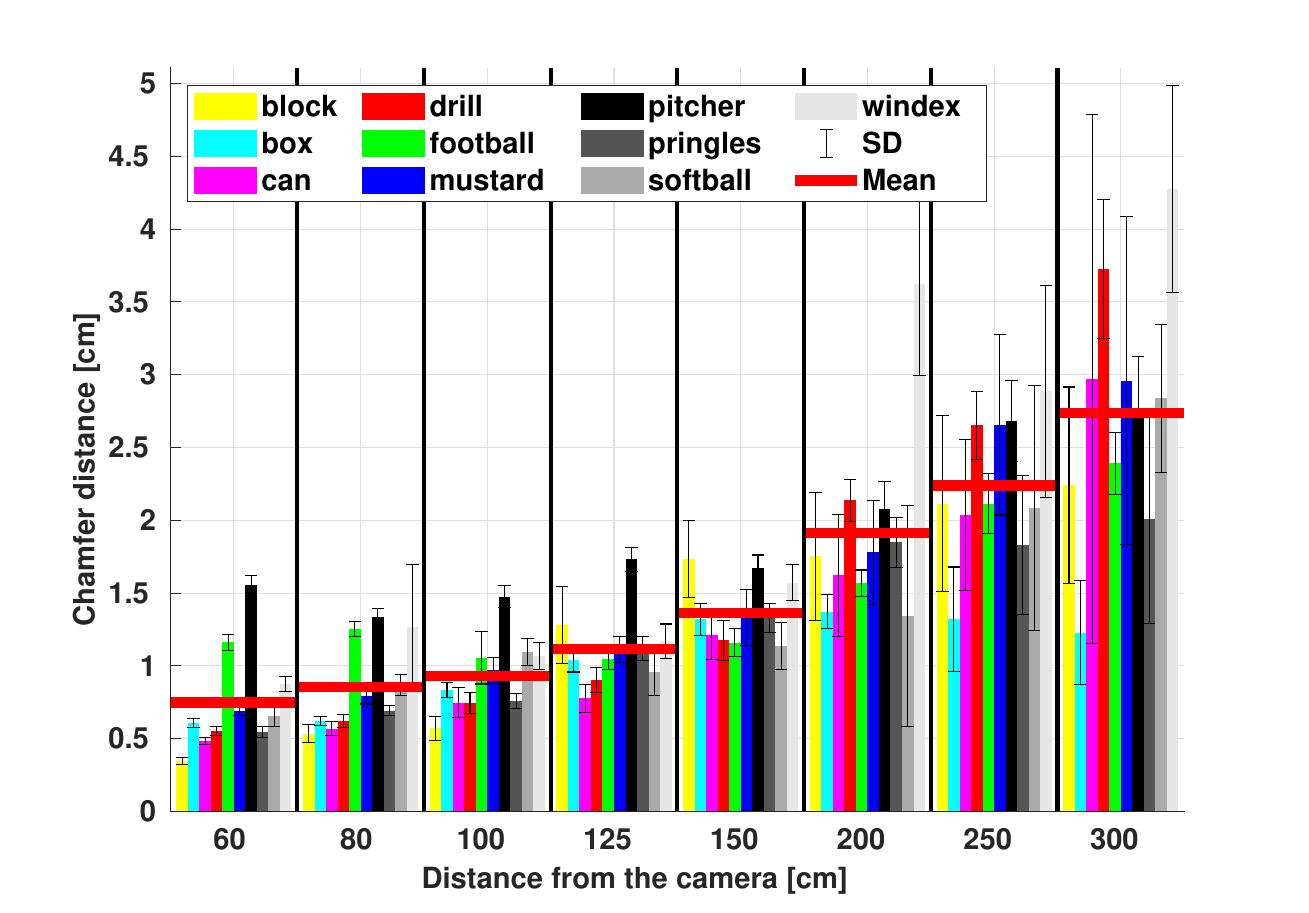}
        (d) OAK-D Pro.
    \end{minipage}
    
    \caption{YCB objects perception -- \acf{cd} for individual objects. Red lines represent mean over all objects for a given distance. Computed over 30 frames for each distance.}
    \label{fig:ycb_cd_per_obj}
\end{figure*}

\figref{fig:ycb_cd_per_obj} provides per-object comparison of \ac{cd}. We provide only \ac{cd}, but all graphs can be found, together with the data used to create them, at \url{https://rustlluk.github.io/rgbd-comparison}. Generally, we can say that the most problematic objects are a \textit{pitcher}, a \textit{windex bottle}, a \textit{football}, and a \textit{drill}---see \figref{fig:objects} for RGB images of the objects. The \textit{windex bottle} is transparent, which is a challenging surface for feature matching. As the bottle was on the table, the cameras could see the table through it, which probably aided perception. If there was nothing behind the bottle, the results would be even worse. The \textit{drill} used has many small features on the surface that are difficult to perceive, mainly at longer distances. In the case of \textit{football} and \textit{pitcher}, the main problem is probably the curvature---and, for the \textit{pitcher}, the handle. We can see that the biggest errors for these two have ZED 2 and OAK-D Pro. It supports the previous results and shows that these two cameras perform better with planar objects and suffer with more complex curved objects.

\textbf{YCB objects perception -- summary.}
Overall, the results using YCB objects correspond to those using the plastic doll. All cameras are comparable up to 150\;cm. After that, the best one is again ZED 2, now with OAK-D Pro being the second best. The reason why OAK-D Pro is not the worst as in the case of doll perception is that the dataset used contains some flat objects that are estimated correctly and decrease the overall error over all the objects. This scenario is the most useful for table-top tasks, such as object perception or grasping. These tasks are usually performed up to a certain range from the robot, and therefore we would advise using D435 as it provides the best price/performance ratio for these tasks.

\section{Conclusion and Discussion}
We compared four state-of-the-art stereoscopic depth cameras on more than 3,000 frames taken with each camera. Segmented data are available at \url{https://rustlluk.github.io/rgbd-comparison}. Using six metrics, we compared the cameras on planar surface perception, plastic doll perception, and on 10 household objects from the YCB dataset. The results show that even though the working principles are similar, individual cameras have different performance in different contexts and at different distances. In conclusion, on the basis of the results, we recommend the following. If your application requires an easy-to-setup and robust solution for lower distances (up to 90-100\;cm) and is more object perception focused with unknown objects of arbitrary shapes, then the choice is RealSense D435. If you want to work at higher distances or also be able to detect planar surfaces properly (e.g., in mobile robots and for \ac{slam}), select RealSense D455 or Stereolabs ZED 2. Here, the choice depends on whether you can afford to have a dedicated \ac{gpu} for depth estimation (ZED 2) or not (D455). If you want to detect planar surfaces and/or objects that are less complex (e.g. boxes), the choice would be ZED 2 or Luxonis OAK-D Pro---again depending on whether you have a \ac{gpu} (then choose ZED 2) or not (choose OAK-D Pro). In case you also require \ac{ai} features, you can choose ZED 2 or OAK-D Pro. The choice here is based on whether you want to compute the \ac{ai} on-board (OAK-D Pro) or have a dedicated computer \ac{gpu} (ZED 2).

It is important to look at all the metrics applicable in a given scenario. In case of planar surface perception, the metrics (bias and standard deviation) are the standard used in the literature~\cite{halmetschlager-funekEmpiricalEvaluationTen2019, heinemann2022MetrologicalApplicationrelatedComparison} and are, to the best of our knowledge, sufficient to properly evaluate the performance. However, this is not the case for the plastic doll and YCB objects perception, as none of the metrics is perfect and sufficient on its own. The \acf{cd} is not robust to incorrect alignment of perceived and ground-truth point clouds, e.g., small rotation of one of the point clouds at the center of mass (mainly for bigger objects) results in high errors as closest points at the boundaries of the object are far away. \acf{js} is robust to these as voxelization can mitigate the effect of small alignment errors. However, it is very sensitive to outliers, e.g., missing handle of a pitcher or hallucinated peaks. $F_1$ score is not informative in cases where there are no outliers. This can be seen in the case of the plastic doll, where the $F_1$ score (\figref{fig:dolls_f}) is not changing until 175\;cm, but for example the \ac{js} (\figref{fig:dolls_jac}) shows decreasing performance. The angle between normals allows to assess performance in the sense of shape similarity but does not capture the effect of outliers and missing points.

The overall best camera is StereoLabs ZED 2. This is a similar finding to \cite{heinemann2022MetrologicalApplicationrelatedComparison}. This camera provided the best results for planar surface estimation in terms of bias and standard deviation. Also, visually the point clouds created are the most flat. For plastic doll and YCB object perception, the camera also showed superior performance. For all metrics, it achieved the best mean performance, while keeping the deviation between individual frames lower than the other cameras. It also provides easily accessible \ac{ai} features such as keypoint detection or face tracking. On the other hand, the camera requires CUDA-enabled \ac{gpu}, which may be a drawback for some applications. ZED 2 also has a long baseline (the distance between two sensors in the physical camera) and works better from distances about 100\;cm, which could be impractical for some tasks such as manipulation. On lower distances it tends to create holes in the object and miss some more complicated parts. From a user point of view, the API and associated software is not as intuitive as for Intel RealSense devices. At first, we had some problems setting the environment and installing the proper versions of the required libraries, which was never the case for RealSense cameras. 

The other cameras have a more specific use case. The Luxonis OAK-D Pro camera provides on-board \ac{ai} features without the need for an external GPU. It works great for planar surface, where it provides performance comparable with ZED 2 camera, while computing everything on-board. However, it's performance decreases a lot when perceiving more complex objects. The camera basically groups points into layers of constant depths. At lower distances, the layers are close to each other, so the overall perceived shape is correct. At higher distances, the layers are far from each other---see \figref{fig:planes}, \figref{fig:dolls}, and \figref{fig:ycb}. The same behavior was discovered in \cite{heinemann2022MetrologicalApplicationrelatedComparison}, where they discarded the camera from the comparison on a spherical object because the errors were too high. The effect is visible in the difference between \ac{js} and \ac{cd}. Performance, for example in plastic doll perception with \ac{js} at higher distances, is comparable to the RealSense devices, but for \ac{cd} the error is notably higher. We speculate that this behavior arises from optimizations in image resolution that allows all computations to be performed on board of the sensor. This makes the camera usable for more articulated objects only at lower distances. For simpler and more planar objects, it can be still used at higher distances and the onboard computations make it usable, for example, in indoor navigation and mapping. In addition, the camera seems to \enquote{load} when started, i.e., sometimes first frames after the start tend to be unfocused and more noisy. The API and software are the least intuitive of all the cameras---as also stated in \cite{heinemann2022MetrologicalApplicationrelatedComparison}.

The RealSense D435 and D455 are the most typical RGB-D cameras currently used in robotics (this is supported by the fact that the camera are chosen in most of the \sotan{} in camera comparison, e.g. \cite{heinemann2022MetrologicalApplicationrelatedComparison, servi2024ComparativeEvaluationIntel, halmetschlager-funekEmpiricalEvaluationTen2019,fu2020ApplicationConsumerRGBDd}). They work on the same stereoscopic principle with IR projection computing everything using on-board CPU without any additional \ac{ai} features. The difference in these is the baseline that results in different performance.  The D435 has a smaller baseline and therefore works at lower distances. The ideal range is only up to 3\;m, so the performance decreases quickly with increasing distance, but up to 100\;cm, the planar estimation is on par with other cameras (except for ZED 2, which is better) and for shape estimation it is overall the best in this range. The D455 seems to work better on the plane estimation task and works better for larger distances---the same was discovered in \cite{servi2024ComparativeEvaluationIntel, heinemann2022MetrologicalApplicationrelatedComparison}. The stereo baseline is an important parameter, as with a fixed baseline the depth estimation error grows quadratically~\cite{Gallup2008_Variable}, which is nicely visible in \figref{fig:bias_precision}. This is true for cameras with \enquote{classic} feature matching. We can see, again in \figref{fig:bias_precision}, that it is not the case for ZED 2 which computes depth values using a neural network. The second parameter influenced by the baseline is the minimum distance in front of the camera---again, it is not true for ZED 2. The bigger the baseline, the larger the necessary distance in front of the camera. This property is a problem mainly for close proximity sensing---see \cite{Docekal_closeProximity}.

The RealSense API and all other software is the best among all the cameras. However, we also encountered some problems. Sometimes, the cameras would not start properly without removing and reinserting the USB cable. Also, mainly D455 seems to have a fixed order of turning on RGB and depth channels. If it is done in incorrect order, the depth and RGB images are not correctly matched. D455 also has a low RGB resolution (1280x800), which can be, mainly together with high FOV, a problem for RGB-based computer vision applications such as keypoint detection.

Light conditions during sensing are important, which is also tied to the fact whether the cameras use active (IR projection) or passive technology. \etal{Halmetschlager-Funek}~\cite{halmetschlager-funekEmpiricalEvaluationTen2019} showed that light intensity influences the results and IR projection can help in darker environments and also with reflective surfaces in brighter environments. Furthermore, in \cite{heinemann2022MetrologicalApplicationrelatedComparison} the authors showed that ZED 2, as a passive camera without IR projection, has shown lower performance under low lighting. Also, when there is direct sunglight, performance is very limited~\cite{neupane2021EvaluationDepthCameras}. We tried to keep the light conditions as consistent as possible. However, we were unable to change them to test different ones. Such experiments could be added in the future. 

Future work should also focus on comparing \acf{tof} sensors. Moreover, the cameras used are the state of the art at the time of publication, but the field is evolving rapidly, and new studies will be necessary when new sensors become available. 

\bibliographystyle{unsrt}
\bibliography{rustl_bib.bib}

\newpage
\begin{IEEEbiography}[{\includegraphics[width=1in,height=1.25in,clip,keepaspectratio]{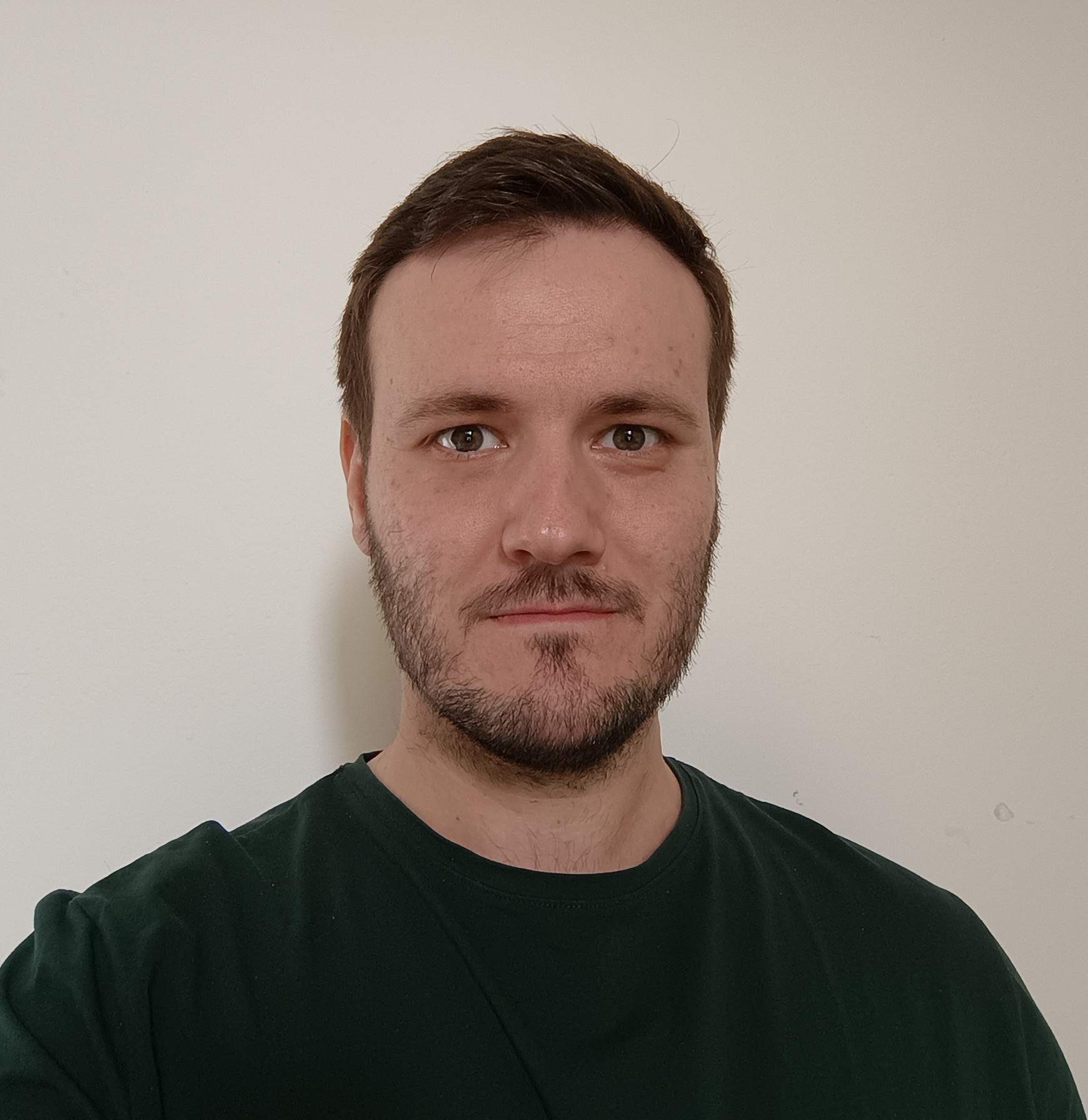}}]{Lukas Rustler} received his BSc. and MSc. degrees in Cybernetics and Robotics from the Faculty of Electrical Engineering, Czech Technical University in Prague in 2019 and 2022, respectively. 

He is a doctoral student at the Humanoid and Cognitive Robotics Group (Faculty of Electrical Engineering, Czech Technical University in Prague) under the supervision of Matej Hoffmann. His research interests include robot kinematic calibration, physical human-robot interaction, grasping, and active multi-modal perception.
\end{IEEEbiography}

\vspace{-20em}

\begin{IEEEbiography}
[{\includegraphics[width=1in,height=1.25in,clip,keepaspectratio]{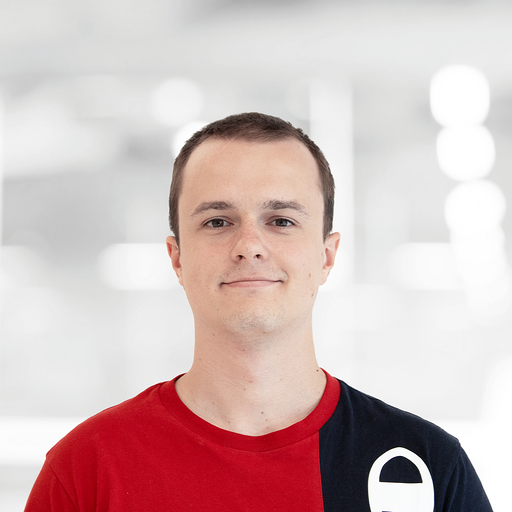}}]{Vojtech Volprecht} received his BSc. and MSc. degrees in Artificial Intelligence from the Faculty of Electrical Engineering, Czech Technical University in Prague in 2022 and 2024, respectively.

He is currently working as a researcher and software developer in the medical company BTL Medical Technologies. His research interests include planning in robotics, perception and estimation from RGB-D data and artificial intelligence in medicine. 
\end{IEEEbiography}

\vspace{-20em}

\begin{IEEEbiography}
[{\includegraphics[width=1in,height=1.25in,clip,keepaspectratio]{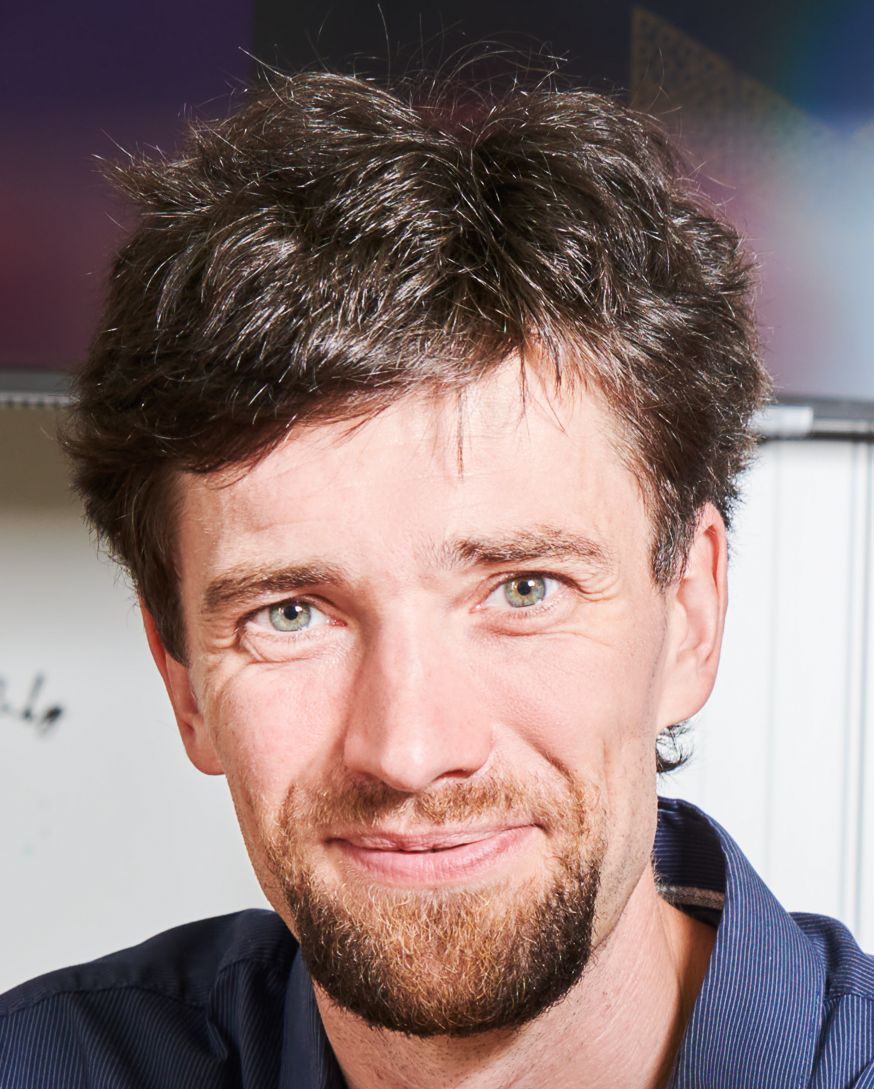}}]{Matej Hoffmann} (Senior Member, IEEE) received the Ph.D. degree in informatics from Artificial Intelligence Lab, University of Zurich, Zurich, Switzerland, in 2012. 

From 2013 to 2016, he conducted postdoctoral research with the iCub Facility of the Italian Institute of Technology, Genoa, Italy, supported by a Marie Curie Intra-European Fellowship. In 2017, he joined the Department of Cybernetics, Faculty of Electrical Engineering, Czech Technical University in Prague, where he is currently an Associate Professor and the Coordinator of the Humanoid and Cognitive Robotics Group. His research interests include humanoid, cognitive developmental, and collaborative robotics, as well as active perception for robot manipulation and grasping.
\end{IEEEbiography}

\EOD

\end{document}